\pdfoutput=1
\documentclass[11pt]{article}

\usepackage[preprint]{acl}
\usepackage{longtable}
\usepackage{placeins}
\usepackage{tabularx}
\usepackage{amssymb}
\usepackage{amsmath}
\usepackage{booktabs}
\usepackage{tablefootnote}
\usepackage{longtable}
\usepackage{subcaption} 
\usepackage{bm}
\usepackage{float}
\usepackage{times}
\usepackage{latexsym}
\usepackage{multirow}
\usepackage[T1]{fontenc}
\usepackage{enumitem}
\usepackage[utf8]{inputenc}
\usepackage{microtype}
\usepackage{inconsolata}
\usepackage{graphicx}
\usepackage{lscape}
\usepackage{nameref}
\title{Influence-driven Curriculum Learning for Pre-training on Limited Data}
\author{
 \textbf{Loris Schoenegger\textsuperscript{1,2}},
 \textbf{Lukas Thoma\textsuperscript{1,2}},\\
 \textbf{Terra Blevins\textsuperscript{1,4}},
 \textbf{Benjamin Roth\textsuperscript{1,3}}
\\
\\
 \textsuperscript{1}Faculty of Computer Science, University of Vienna, Vienna, Austria\\
 \textsuperscript{2}UniVie Doctoral School Computer Science, University of Vienna, Vienna, Austria\\
  \textsuperscript{3}Faculty of Philological and Cultural Studies, University of Vienna, Vienna, Austria\\
 \textsuperscript{4}Khoury College of Computer Sciences, Northeastern University, Boston, USA\\
\\
 \small{
   \textbf{Correspondence:} \href{mailto:loris.schoenegger@univie.ac.at}{loris.schoenegger@univie.ac.at}
 }
}
\begin{document}

\maketitle
\begin{abstract}
Curriculum learning, a training technique where data is presented to the model in order of example difficulty (e.g., from simpler to more complex documents), has shown limited success for pre-training language models.
In this work, we investigate whether curriculum learning becomes competitive if we replace conventional human-centered difficulty metrics with one that more closely corresponds to example difficulty as observed during model training.
Specifically, we experiment with sorting training examples by their \textit{training data influence}, a score which estimates the effect of individual training examples on the model's output.
Models trained on our curricula are able to outperform ones trained in random order by over 10 percentage points in benchmarks, confirming that curriculum learning is beneficial for language model pre-training, as long as a more model-centric notion of difficulty is adopted. 
\end{abstract}
\section{Introduction}
Curriculum learning, a training paradigm where the training data is presented to the model in non-random order \cite{bengio_curriculum_2009}, has recently been explored extensively as a pretraining strategy for language models due to its potential to improve performance in low-resource settings \cite{timiryasov_baby_2023}, reduce training time \cite{platanios_competence-based_2019}, or to make the training process more data-efficient and developmentally plausible (i.e., more similar to how humans acquire language; \citealp{warstadt_call_2023,hu_findings_2024}).
A popular form of curriculum learning relies on heuristics that \textbf{sort training data by increasing difficulty} (e.g., lexical diversity trough type-token ratio: \citealp{mi_mmi01_2023}).
However, in low-resource language modeling, approaches that incorporate this curriculum learning strategy have not yielded the anticipated improvements and show no consistent positive effect on model performance \cite{hu_findings_2024}.
In this work, we therefore investigate whether curriculum learning becomes competitive for language model pretraining, if we replace human-centered difficulty measures with one that better reflects training dynamics. 
Specifically, we derive a novel form of curriculum from \textbf{training data influence estimates}, that we obtain from a surrogate model trained with randomly ordered data: These estimates assign documents from the training data scores proportional to their impact on the model's output. We adapt a \textit{gradient similarity-based} influence score \cite{pruthi_estimating_2020}, where influence is measured by comparing loss-gradients of training and test instances, with higher similarity signifying greater influence.
\begin{figure}[ht]
\centering
  \includegraphics[width=0.85\columnwidth, trim=0 0 0 0, clip]{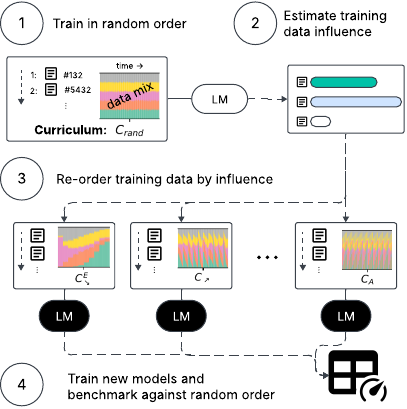}
  \caption {In our method, we extract training data influence estimates from models trained in random order, to create better-performing curricula.}
\end{figure}
We experiment with 10 different sorting strategies, all based on the \textbf{average influence} that a given training example exerts on the prediction of \textit{other} examples sampled from the training data.
We compare model performance under these curricula to both random training and curriculum learning using three human-centered difficulty heuristics.
Through experiments with RoBERTa- \cite{liu_roberta_2019} and Llama models \cite{touvron_llama_2023}, we demonstrate that our approach is more effective than handcrafted curricula, and analyze what ranking and coverage strategies are most effective.
We find that source-difficulty curricula, a popular human-centered design that arranges datasets by their difficulty, are ineffective compared to alternative dataset coverage strategies, and we offer insights into the reasons for their low performance.
Our \textbf{main contributions} are as follows:\footnote{We release our code at \url{https://doi.org/10.5281/zenodo.16919045}, and host all datasets and models on the Hugging Face Hub at \url{https://huggingface.co/collections/loris3/ticl-68a6fd8bcc3093f239439e42}.} 
\begin{enumerate}[nosep, topsep=0pt,label=(\arabic*)]
    \item We demonstrate that our curricula yield an increase of over 10 percentage points (pp) in accuracy for RoBERTa- and over 4 pp for Llama models on a popular challenge dataset for low-resource pre-training (BabyLM 10M-word dataset: \citealp{choshen_call_2024}).
    \item We analyze the data mix of the generated curricula (e.g., child-directed speech, dialogue, etc.) and how it evolves over time;
    \item Analyze loss trajectories to study how our curricula affect the model's learning process;
    \item Explore how example ordering within influence curricula relates to existing heuristics.
\end{enumerate}
\section{Related Work}\label{related-work}
\paragraph{Curriculum Learning} can roughly be categorized into dynamic and static approaches. Dynamic designs incorporate difficulty heuristics directly into the training process, generating or updating the curriculum during training (e.g., \citealp{kumar_self-paced_2010,sedova_learning_2023}).
Static curricula have recently proven popular in the \textit{BabyLM challenge}, a competition promoting the creation of more developmentally plausible language models \cite{hu_findings_2024}: 
Motivated by the observation that humans only require up to 100 million words to reach native levels in a language \cite{gilkerson_mapping_2017}, this challenge invites NLP researchers to explore human-centered learning strategies on a dataset of just 10M or 100M words. 
Participants have incorporated various sorting heuristics into curriculum learning schemes, such as sorting by increasing sentence length \cite{platanios_competence-based_2019,ghanizadeh_towards_2024,borazjanizadeh_optimizing_2023,spitkovsky_baby_2010},
document- or sentence complexity \cite{oba_babylm_2023,opper_effect_2023}, lexical diversity \cite{mi_mmi01_2023,ghanizadeh_towards_2024}, or dataset-level source difficulty by category \cite{thoma_cogmemlm_2023,
huebner_babyberta_2021, martinez_climb_2023,opper_effect_2023}.
However, static approaches following this framework have shown no consistent positive effect on model performance \cite{hu_findings_2024}.

Our method is motivated by the assumption that children's language learning proceeds from easy to complex input \cite{elman_learning_1993}, but represents a middle ground between static and dynamic approaches: we generate static curricula, but base them on a score that reflects training dynamics.
\paragraph{Training Data Influence for CL}
\citeauthor{bejan_make_2023} (\citeyear{bejan_make_2023}) employ TracIn self-influence \cite{pruthi_estimating_2020} for curriculum learning in the fine-tuning setting. For them, self-influence is defined as $\nabla \ell(w_{t},z) \cdot \nabla \ell(w_{t},z)$ \cite{pruthi_estimating_2020}, which does not relate to other data points in the training data, and effectively only quantifies magnitude for a given example.
In contrast to our approach, their focus lies on improving performance by filtering outliers and up-weighting the most influential examples.
Our approach incorporates more information, specifically pairwise influence scores between one example and \textit{all} other examples in the training data, as outlined in Section \ref{tdm}. 
\paragraph{Role of Example Difficulty in Learning}
Several authors have utilized measures of example difficulty to systematically study the effect of curriculum learning for supervised fine-tuning tasks and in the image domain 
\cite{hacohen_power_2019,wu_when_2020, jiang_characterizing_2021, baldock_deep_2021}.
For instance, \citeauthor{wu_when_2020} (\citeyear{wu_when_2020}), study whether examples of similar difficulty are learned at similar stages across architectures through comparing the \textit{learned iteration} of examples across models, a metric defined as the first epoch at which the model correctly predicts them. Our setup differs in that we study the model's downstream performance and operate within an unsupervised setting.
\section{Methodology}\label{methodology}
In this work, we investigate the benefits of incorporating training data influence estimates into curriculum learning methods, particularly for low-data pretraining settings. We first introduce our approach for estimating example difficulty using training gradients. Then, we describe our curriculum designs and outline our experimental setup.
\subsection{Training Data Influence Estimation}\label{tdm}
\begin{figure}[ht]
  \includegraphics[width=\columnwidth]{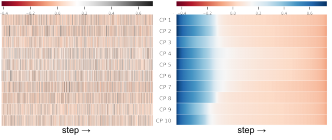}
  \caption{Left: measured influence on $C_{rand}$; Right: anticipated influence if sorted according to $C_{\searrow}$.}
  \label{influence-matrix-example}
\end{figure}

We define a new metric for measuring example difficulty in curriculum design that leverages training data influence estimates: We adapt TracinCP \cite{pruthi_estimating_2020} for this, which in its original formulation estimates the \textit{point-wise influence} $\phi_\text{TracInCP}(z,z')$ that training on an instance $z$ had on the model, when predicting a test instance $z'$.
The estimation process involves measuring the similarity between the gradients of the model's loss function, when evaluated on $z$ and $z'$ respectively, w.r.t some set of parameters $w_{t}$, and is repeated at a series of checkpoints $T$:
\begin{equation}
\resizebox{0.895\linewidth}{!}{$
        \phi_\text{TracInCP}(z,z') = \displaystyle\sum_{\forall t \in T} \eta_t \nabla \ell(w_{t},z) \cdot \nabla \ell(w_{t},z') 
        $}
\end{equation}
Following \citet{yeh_first_2022}, we let $w_{t}$ be the model's input embeddings at checkpoint $t$.\footnote{Note that this score incorporates information about the full model, as the gradient chains through higher layers as well \cite{yeh_first_2022}.}
To leverage this point-wise influence score \textbf{for curriculum learning}, we propose to calculate the \textbf{average influence} $\phi_t(z, D)$ that a given training example exerts on the prediction of all other examples from the training data $D$. Omitting the learning rate $\eta_t$, for one training instance $z$, and one checkpoint $t$ we calculate:
\begin{align}
    \phi_t(z, D) 
    &= \frac{\sum_{\forall z' \in D} \nabla\ell(w_{t},z) \cdot \nabla\ell(w_{t},z')}{|D|}   \\ 
    &= \nabla\ell(w_{t},z) \cdot \mathbb{E}_{z' \sim D} [\nabla\ell(w_{t},z')]
\end{align}
Intuitively, this score quantifies the average utility of a given example during training. Unlike measures of surprisal such as perplexity, it is high for prototypical examples (which feature loss gradients similar to the average gradient) and low for outliers.
Doing so for all examples in the training dataset $D$, at regular checkpoints for a model trained in random order, yields a matrix $\Phi \in 	\mathbb{R}^{|D| \times |T|}$ like the one depicted in Figure \ref{influence-matrix-example}, which we subsequently use for constructing curricula with various reordering functions.
In initial experiments, we observed that this score based on dot-product similarity was biased against longer examples, which was also observed by \citet{xia_less_2024}. Thus, \textbf{we normalize the loss gradients} to reduce the impact of gradient magnitude on the similarity scores, effectively yielding cosine similarity \cite{hammoudeh_identifying_2022,hammoudeh_training_2024,park_trak_2023,xia_less_2024}. 
\subsection{Curriculum Design}
This section introduces our 10 curriculum design methods based on influence estimates, as well as 4 baseline curricula. Our designs can be broadly categorized into two categories, characterized by their coverage strategy: the first group of curricula covers the full dataset every epoch, while the second group progressively increases example difficulty across epochs, consequently not re-visiting examples from early epochs in later ones. 
\paragraph{Epoch-wise Dataset Coverage Strategies}\mbox{}\\
In the curricula $C_\searrow$ and $C_\nearrow$, we sort documents in descending ($\searrow$) or ascending ($\nearrow$) order of influence, measured using model checkpoints of a surrogate model trained in random order stored after each epoch $t$.
We include an additional pair of curricula $C_\searrow^\sim$ and $C_\nearrow^\sim$, where, in an attempt to increase data diversity during training, we additionally divide the curriculum into ordered subsets of 1000 documents, and then randomly shuffle the documents within these subsets.
Similarly, motivated by the intuition that examples with lasting influence across epochs should be prioritized because they appear to have been more difficult for the surrogate model to learn, we add a re-weighting step to the two curricula $(C*h)_\searrow^\sim$ and $(C*h)_\nearrow^\sim$, where we convolve the influence estimates $\Phi$ with a lognormal filter $h$ before the sorting step; this thus up-weights examples that remain influential in subsequent epochs: $(C*h)_{(t,i)} = \sum_{k=0}^{T}  \Phi_{(t-k,i)}\cdot \text{h}(k)$.

Lastly, emulating prior works that used influence estimates for data cleaning and not solely for re-ordering (e.g., \citealp{bejan_make_2023}), we add a curriculum $C^{\{50\}}$ where we discard the 50\% least influential examples in each epoch, while keeping the total number of words shown to the model constant. We shuffle once per epoch. 
\paragraph{Cumulative Dataset Coverage Strategies}\mbox{}\\
Source difficulty curricula \cite{martinez_climb_2023} are a curriculum learning strategy where models are trained on a collection of datasets that are manually sorted by difficulty (but the individual examples within these datasets are not).
In $C^E_\searrow$ and $C^E_\nearrow$, we design a similar coverage strategy, allowing us to subsequently test whether curricula based on training data influence yield similar dataset mixtures as handcrafted ones: 
In contrast to the curriculum designs introduced so far, we aggregate the individual influence estimates for a given example across all $T$ epochs to obtain a measure of its overall influence during training ($\phi_T(z,D) = \sum_{\forall t}\phi_t(z,D)$). We then sort examples by this score, either in ascending ($\nearrow$) or descending order ($\searrow$). Subsequently, we divide this ordered data into $m=10$ segments, from which we then randomly sample to create $m$ equal-length epochs with examples of increasing or decreasing difficulty respectively.

Our last curriculum, $C_A$, is designed as a compromise between curricula with epoch-wise dataset coverage strategies and $C^E_\cdot$: 
In this curriculum, we alternate between showing subsets of high influence scores and subsets of low influence scores, but shuffle the individual examples within each segment randomly. Specifically, we first sort examples by their aggregate score $\phi_T(z,D)$, and create $m=10$ segments just as for $C^E_\cdot$. We then assemble the curriculum from these segments by alternating between the highest-influence and lowest-influence ones until all are used. We train for 10 epochs in this order, randomly shuffling the examples within each segment before each pass.
\paragraph{Baseline Curricula}\mbox{}\\
We include 4 baseline curricula $C_{rand}$, $C_{source}$, $C_{MATTR}$ and $C_{PPL}$: In $C_{rand}$ we emulate non-curriculum learning, performing 10 full passes over the training data in random order. We train one model per dataset using this curriculum, storing checkpoints after each full pass so that it can serve both as a surrogate model for extracting influence estimates and as a baseline (we utilize a total of $T=10$ checkpoints).

Handcrafted \textit{source-difficulty curricula} present datasets sorted by difficulty as distinct blocks (e.g., children's books before Wikipedia articles). We define such a curriculum in $C_{source}$, by assigning the datasets in Table \ref{tab:source-difficulty-combined} to one of 5 stages (C1-C5), following previous work \cite{thoma_cogmemlm_2023,huebner_babyberta_2021, martinez_climb_2023,opper_effect_2023}. Similar to $C^E_\nearrow$ and $C^E_\searrow$, we train for two epochs per stage, randomly shuffling examples within each epoch.

$C_{MATTR}$ is inspired by \citeauthor{mi_mmi01_2023}'s (\citeyear{mi_mmi01_2023}) use of \textit{type-token ratio} (TTR) for curriculum learning. Here, we sort documents by increasing moving average type-token ratio ($C_{MATTR}$ (with a window length of 5); \citealp{covington_cutting_2010}).\footnote{We choose to use MATTR over TTR as a metric to make our curricula more robust to variation in document length.} Lastly, for $C_{PPL}$, we sort in order of increasing perplexity under a static uni-gram model, as described in \citet{martinez_climb_2023}. With both $C_{MATTR}$ and $C_{PPL}$, we train the model on full epochs in this order 10 times.
\begin{table*}[ht]
\centering
\scriptsize
\begin{subtable}[t]{0.4\textwidth}
    \centering
    \begin{tabular}{l p{0.8\columnwidth}}
\hline
\textbf{C1} & \textbf{Child Directed Speech} \\ 
 & CHILDES \cite{macwhinney_childes_2014} \\ \hline
\textbf{C2} & \textbf{Unscripted Dialogue} \\ 
 & Switchboard Dialog Act Corpus \cite{stolcke_dialogue_2000} \\ 
 & British National Corpus (BNC), \newline dialogue portion \cite{consortium_british_2007}\\ \hline
\textbf{C3} & \textbf{Scripted Dialogue} \\ 
 & OpenSubtitles \cite{lison_opensubtitles2016_2016} \\ \hline
\textbf{C4} & \textbf{Wiki} \\ 
 & Simple Wiki \cite{warstadt_call_2023} \\ \hline
\textbf{C5} & \textbf{Written English} \\ 
 & Standardized Project Gutenberg Corpus \cite{gerlach_standardized_2018} \\ \hline
\end{tabular}
    \label{source-difficulty-2024}
\end{subtable}
\hspace{0.5cm}
\begin{subtable}[t]{0.5\textwidth}
    \centering
    \begin{tabular}{l p{0.9\columnwidth}}
\hline
\textbf{C1} & \textbf{Child Directed Speech} \\ 
 & CHILDES \cite{macwhinney_childes_2014} \\ \hline

\textbf{C2} & \textbf{Children's Books} \\ 
 & Children Stories Text Corpus \cite{bensaid_fairytailor_2021}\\
 & Children's Book Test \cite{hill_goldilocks_2016}\\  \hline

\textbf{C3} & \textbf{Dialogue} \\ 
 & Switchboard Dialog Act Corpus \cite{stolcke_dialogue_2000} \\ 
 & British National Corpus (BNC), dialogue portion \cite{consortium_british_2007}\\ 
 & OpenSubtitles \cite{lison_opensubtitles2016_2016} \\ \hline

\textbf{C4} & \textbf{Educational} \\ 
 & Simple Wiki \cite{warstadt_call_2023} \\ 
 & QED \cite{abdelali_amara_2014} \\ \hline
\textbf{C5} & \textbf{Written English} \\ 
 & Standardized Project Gutenberg Corpus \cite{gerlach_standardized_2018} \\ 
 & Wikipedia \cite{warstadt_call_2023} \\ \hline
\end{tabular}
    \label{source-difficulty-stratified-equitoken}
\end{subtable}
\caption{Curriculum stages in $C_{source}$. Stages for $D_{2024}$ (left) differ from those in $D_{stratified}$ and $D_{equitoken}$ (right) to allow for a balanced split. We make all three datasets available under CC BY 4.0.}
\label{tab:source-difficulty-combined}
\end{table*}
\subsection{Datasets}\label{datasets}
We train models on three datasets: 
\begin{itemize}[leftmargin=*, itemsep=0em]
    \item $D_{2024}$ is the 10M word text-only dataset utilized in the 2024 and 2025 iterations of the BabyLM challenge \cite{choshen_call_2024,charpentier_babylm_2025}, which is composed of datasets of various levels of difficulty listed in Table \ref{tab:source-difficulty-combined}. 
    \item To facilitate analysis of source-difficulty curricula, we construct $D_{stratified}$, which has an equal number of words per stage. We sample from the same datasets underlying $D_{2024}$, but add sources to balance word counts (Table \ref{tab:source-difficulty-combined}).
    \item As document length varies substantially by source, we additionally control for the number of words per document in a third dataset $D_{equitoken}$ (also stratified and balanced w.r.t stages); specifically, we create synthetic documents that are exactly 100 words long by concatenation.
\end{itemize}
\noindent Finally, we create a shared evaluation set for all $D_*$, sampled from the 100M word version of said BabyLM dataset ($|D_{eval}| = 0.05 \cdot |D_{2024}|$). 
\subsection{Models}\label{models}
Our experiments produce a total of 84 models, one RoBERTa- (126M params) and one Llama model (97.2M params), both with random initializations, for each combination of the 3 datasets and 14 curricula. 
We train on 4 NVIDIA H100 GPUs with an effective batch size of 2048, using the parameters summarized in Table \ref{training-params} in Appendix \ref{appendix-implementation-details}.
Each curriculum includes at most 100 million words (e.g., 10 passes over a dataset of 10M tokens for $C_{rand}$).
\section{Results and Analysis}
\begin{figure*}[t]
\centering
  \includegraphics[width=0.85\textwidth, trim=0cm 0.35cm 0cm 0.35cm, clip]{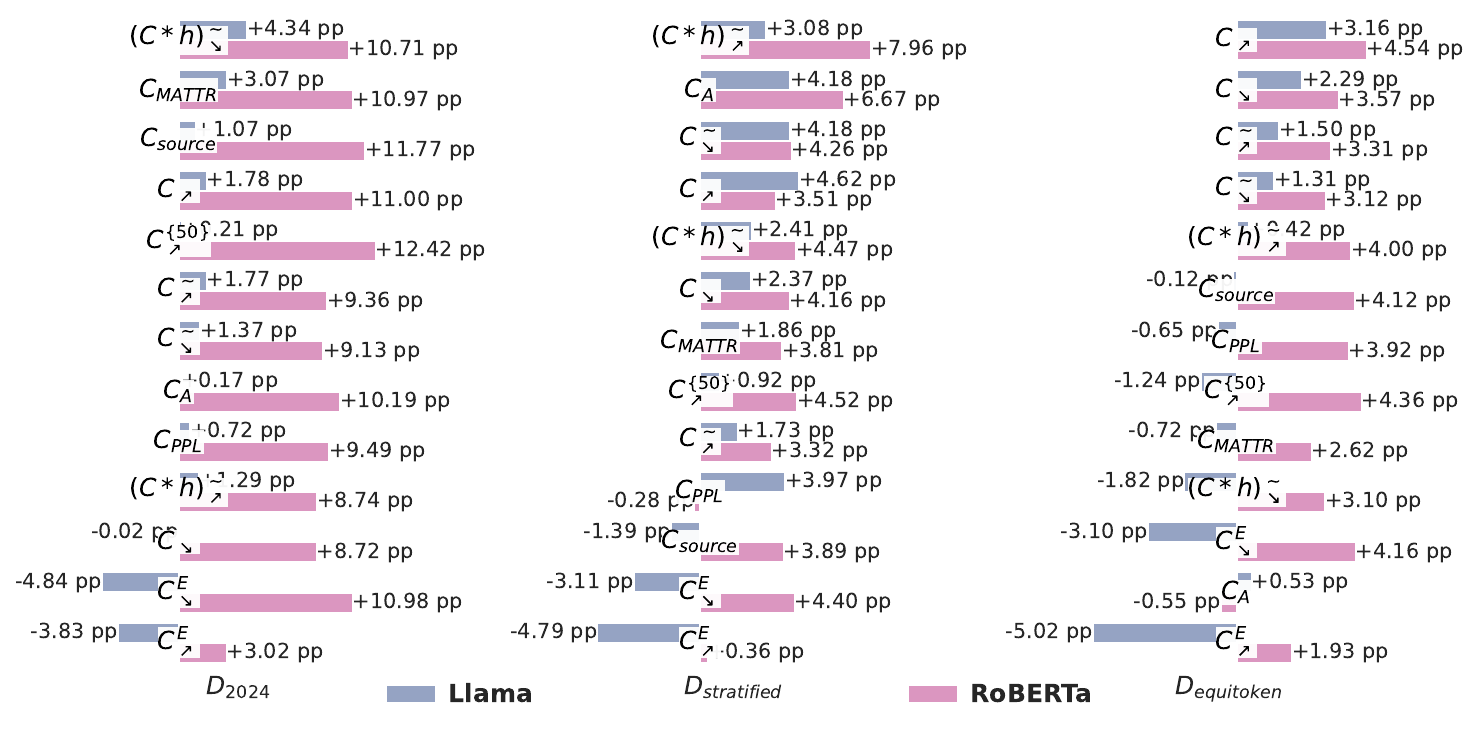}
  \caption{Average change in macro-accuracy across benchmark tasks w.r.t. training on the random curriculum. Sorted by average change across RoBERTa and Llama models.}
  \label{accuracy-delta-per-curriculum}
\end{figure*}
This section presents and analyzes the results of our curriculum design experiments. Specifically, we: (1) present the benchmark performance of our models on downstream tasks; (2) compare the source composition of our curricula to those of the baselines;
(3) analyze training- and evaluation loss trajectories;
(4) and explore how example ordering in the influence curricula correlates with the orderings of existing heuristics.
\subsection{Benchmark Performance}\label{benchmark-performance}
\begin{table}[ht]
\scriptsize
  \centering
\begin{tabular}{lrrrrr}
\toprule
 & Llama & Llama &RoBERTa & RoBERTa\\
Dataset &$C_{rand}$ & $(C*h)^\sim_\nearrow$ & $C_{rand}$ & $(C*h)^\sim_\nearrow$ \\
\midrule
$D_{2024}$ & 0.541 & \textbf{0.554} & 0.466 & 0.553\\
$D_{stratified}$ & 0.536 & 0.566 & 0.512 & \textbf{0.592}\\
$D_{equitoken}$ & 0.523 & 0.527 & 0.492 & \textbf{0.532} \\
\bottomrule
\end{tabular}

  \caption{Macro-Accuracy across tasks for random order and the $(C*h)^\sim_\nearrow$ curriculum. 
  }
    \label{results-random}
\end{table}
\noindent We evaluate our curricula by comparing their performance to models trained on the same data in random order. In Figure \ref{accuracy-delta-per-curriculum}, we report the increase or decrease in macro-accuracy across individual benchmark tasks from BLiMP \cite{warstadt_blimp_2020}, BLiMP supplement \cite{warstadt_findings_2023}, EWOK \cite{ivanova_elements_2024}, Super GLUE \cite{wang_glue_2019},
as well as an entity tracking task \cite{kim_entity_2023} and an adjective nominalization task \cite{hofmann_derivational_2024}, as implemented specifically for the BabyLM challenge \cite{charpentier_babylm_2025}. 
Results for the individual benchmarks are provided in Appendix \ref{appendix-benchmark-results-loss-trajectories}.

In terms of raw performance, the RoBERTa model trained using $(C*h)^\sim_\nearrow$ (sorted by increasing influence, re-weighted with lognormal filter) on $D_{stratified}$ is the \textbf{best performing model overall} (0.592 macro-acc, +7.96 pp over $C_{rand}$), with the best Llama model being the one trained with $(C*h)^\sim_\searrow$ (sorted by decreasing influence) on $D_{2024}$ (0.584, +4.34 pp).
RoBERTa models see higher absolute gains through the addition of curriculum learning than Llama models in our experiments. This can partially be attributed to their lower initial accuracy when trained in random order, with Llama models outperforming RoBERTa models by 7.5, 2.4, and 3.1 pp on the $D_{2024}$, $D_{stratified}$, and $D_{equitoken}$ datasets, respectively (Table \ref{results-random}).
Notably, for RoBERTa models, the handcrafted source curriculum was effective on $D_{2024}$ (+11.77 pp), and only two curricula lead to a decrease in performance, namely $C_{PPL}$ on $D_{stratified}$ (-0.28 pp), and $C_{A}$ on $D_{equitoken}$ (-0.55 pp).
For Llama models, in contrast, the worst-performing curricula $C^E_\searrow$ and $C^E_\nearrow$ incur a considerable 3.10-5.02 pp decrease in accuracy over training in random order. 

For both model architectures, the highest gains through curriculum learning are on $D_{2024}$ followed by $D_{stratified}$ (equal number of words per stage), and $D_{equitoken}$ (equal number of documents per stage, and words per document).
\paragraph{Dataset Coverage Strategies}\mbox{}\\
Models trained with handcrafted- ($C_{source}$) and synthetic source difficulty curricula ($C^E_\searrow$, $C^E_\nearrow$), both designed to increase difficulty gradually across epochs (cumulative coverage strategies), perform worse overall than the other designs, which perform one full pass over the data each epoch (per-epoch coverage strategies).
$C_A$, where we alternate between showing subsets of high influence scores and subsets of low influence scores, shows significant improvements over training in random order for both Llama (+4.18 pp) and RoBERTa (+6.67 pp) on $D_{stratified}$ and $D_{2024}$ for RoBERTa (+10.19 pp), but not for the remaining three models.  
\paragraph{Sorting Direction and Shuffling Strategy}\mbox{}\\
Surprisingly, our benchmark results do not conclusively show whether curricula sorted by ascending ($\nearrow$) or descending ($\searrow$) influence perform better; the ascending version of the same strategy does not consistently outperform the descending version (and vice versa). 
Curricula where we shuffle within stages (e.g., $C^\sim_\searrow$) similarly do not reliably outperform ones without, the same applies to curricula built from lognorm-filtered influence estimates ($(C*h)^\sim_\searrow$). We offer a potential explanation for this in Section \ref{discussion}.
\subsection{Source Composition}\label{datamixanalysis}
\begin{figure*}[ht]
  \includegraphics[width=\textwidth,trim=0cm 0.3cm 0cm 0.25cm, clip]{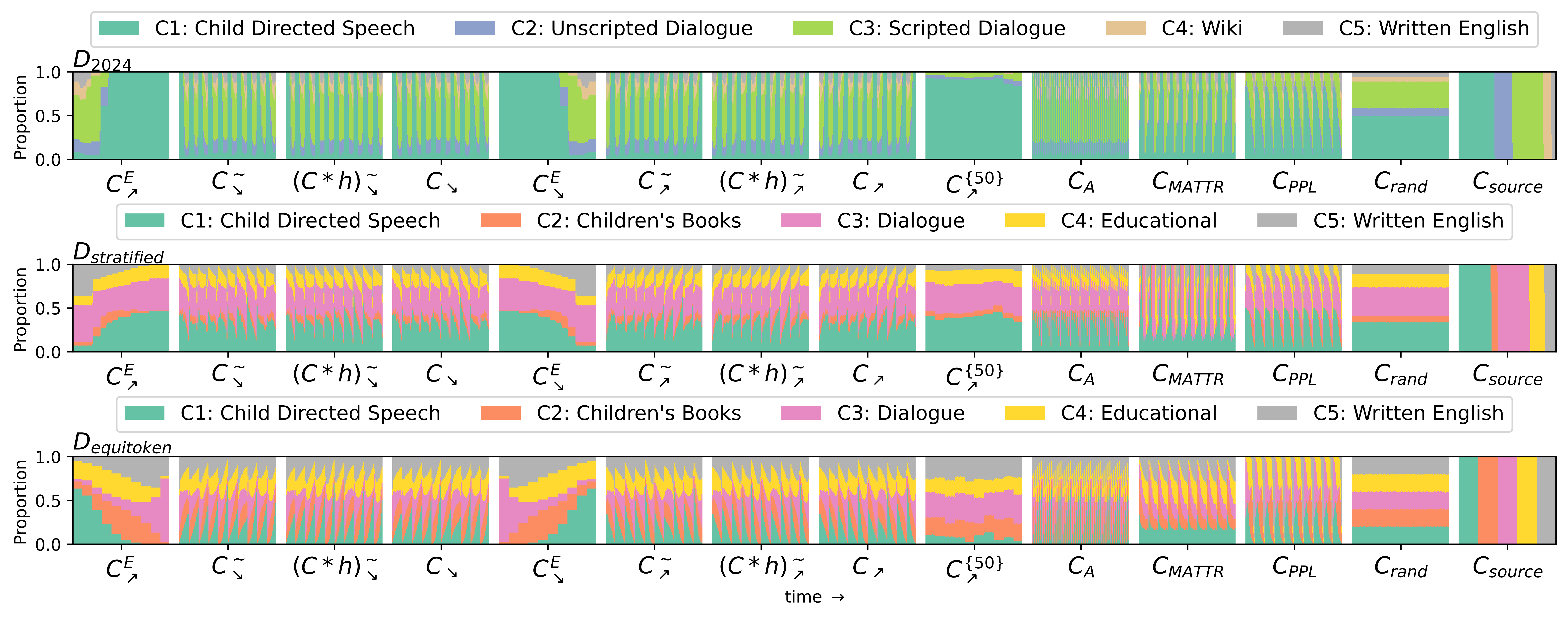}
  \caption{Dataset mix of curricula for Llama models. We trace back documents to the stages defined in Table \ref{tab:source-difficulty-combined}. }
  \label{influence-plosst}
\end{figure*}
The datasets we utilize are themselves composed of sources of varying difficulty; similar to previous work \cite{thoma_cogmemlm_2023} we have attributed each to one of five stages of increasing difficulty (C1-C5; from a human learning perspective) for constructing the handcrafted curricula (Table \ref{tab:source-difficulty-combined}).
Based on these labels, we plot the source compositions of the training data shown to the Llama models over time in Figure \ref{influence-plosst} and provide those of RoBERTa models in Appendix \ref{appendix-complementary-figures}.

We observe that our \textbf{influence curricula are highly sensitive to the source distribution of the dataset}. 
\textit{C1: Child Directed Speech} and \textit{C3: Dialogue}, the two largest stages in the unbalanced $D_{2024}$ dataset, are scheduled first in the synthetic source difficulty curriculum $C^E_\searrow$, with more than half of the training steps allocated to them.
For $C^{\{50\}}$, where we discard the 50\% least influential examples in each epoch, the share of child directed speech accounts for over 90\% of examples throughout the training process, despite accounting for only roughly half of $D_{2024}$ by number of documents.

This \textbf{over-representation of child directed speech} in the majority of epochs may explain why these curricula perform worse in benchmark tasks than all other influence curricula across all datasets and model types:
When controlling for the number of words per source ($D_{stratified}$), the effect is less extreme, yet, \textit{C1: Child-Directed Speech},  \textit{C3: Dialogue}, and \textit{C4: Educational} are more frequently shown in early rather than in later epochs in $C^E_\searrow$, with \textit{C5: Written English} following the opposite trend. 
For $D_{equitoken}$ however, where the model used for influence estimation sees an equal number of tokens and documents per stage, all trends are reversed, with C1 now shown more often in later epochs, and C5 in earlier ones. 
One possible explanation stems from the definition of our datasets, which sample based on a word-based budget rather than one based on the number of documents:
In $D_{2024}$, C1 accounts for 54\% of documents but only 28\% of words, while C5 comprises 25\% of words within just 6\% of the dataset's documents.\footnote{the same pattern applies in $D_{stratified}$}
Because our sorting relies on a \textit{per-document average} influence measure, similarity to the larger subset C1 likely disproportionately impacts influence scores compared to similarity with C5. This suggests that our ranking method is \textbf{biased against smaller sources} (by number of documents). 

\begin{figure}[ht]
\centering
  \includegraphics[width=0.85\columnwidth, trim=0cm 0.3cm 0cm 0.3cm, clip]{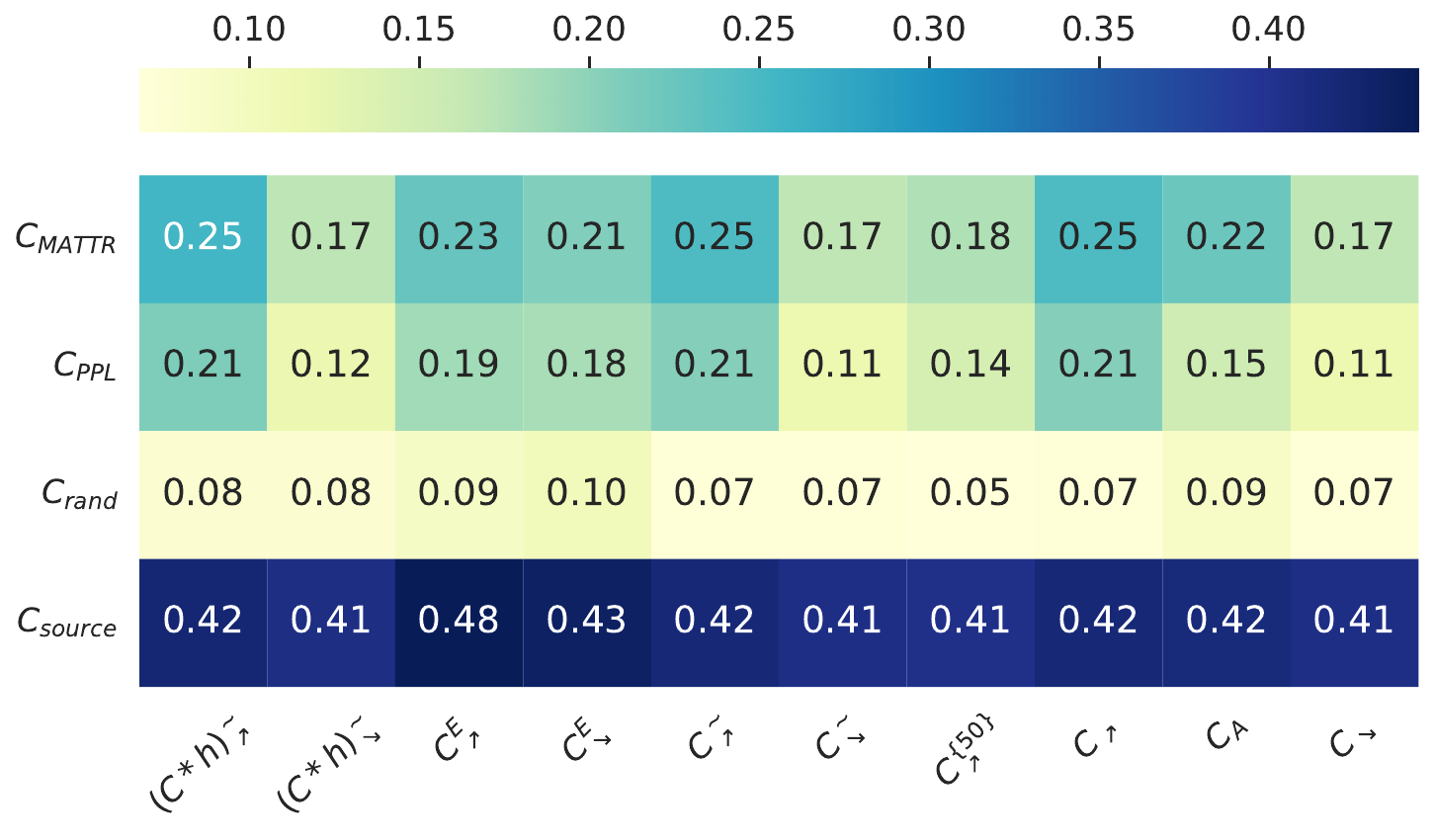}
  \caption{Average Jensen-Shannon divergence between curricula for Llama models. Lower values indicate more similar stage distributions.}
  \label{jensen-shannon-plot-llama}
\end{figure}

Contrary to our initial expectation that the influence of child-directed speech would diminish in later epochs, the \textbf{source composition} of epoch-wise dataset coverage strategies (e.g., $C_\searrow$), \textbf{does not strongly vary over time}.
To obtain a formal measure of how similar a curriculum's source distribution over time is to the model-agnostic baselines, we split both curricula into $n=1000$ segments, for which we then calculate the average Jensen-Shannon divergence\footnote{$\mu_{JSD}(p_a||p_b) :=\sum_{i=1}^{n}\frac{D_{KL}(p^i_a||p^i_b) + D_{KL}(p^i_b||p^i_a)}{2} / {n}$, where $p^i_a$ and $p^i_b$ are the source distributions of two segments}. We find that our curricula's source distribution is closer to that of $C_{rand}$ than to other baselines (i.e., our curricula retain the dataset's source distribution, Figure \ref{jensen-shannon-plot-llama}).
We therefore cannot explain the performance of influence curricula through their source distributions alone. 

\begin{figure*}[ht]
\centering
  \includegraphics[width=0.9\textwidth, trim=0cm 0.3cm 0cm 0.25cm, clip]{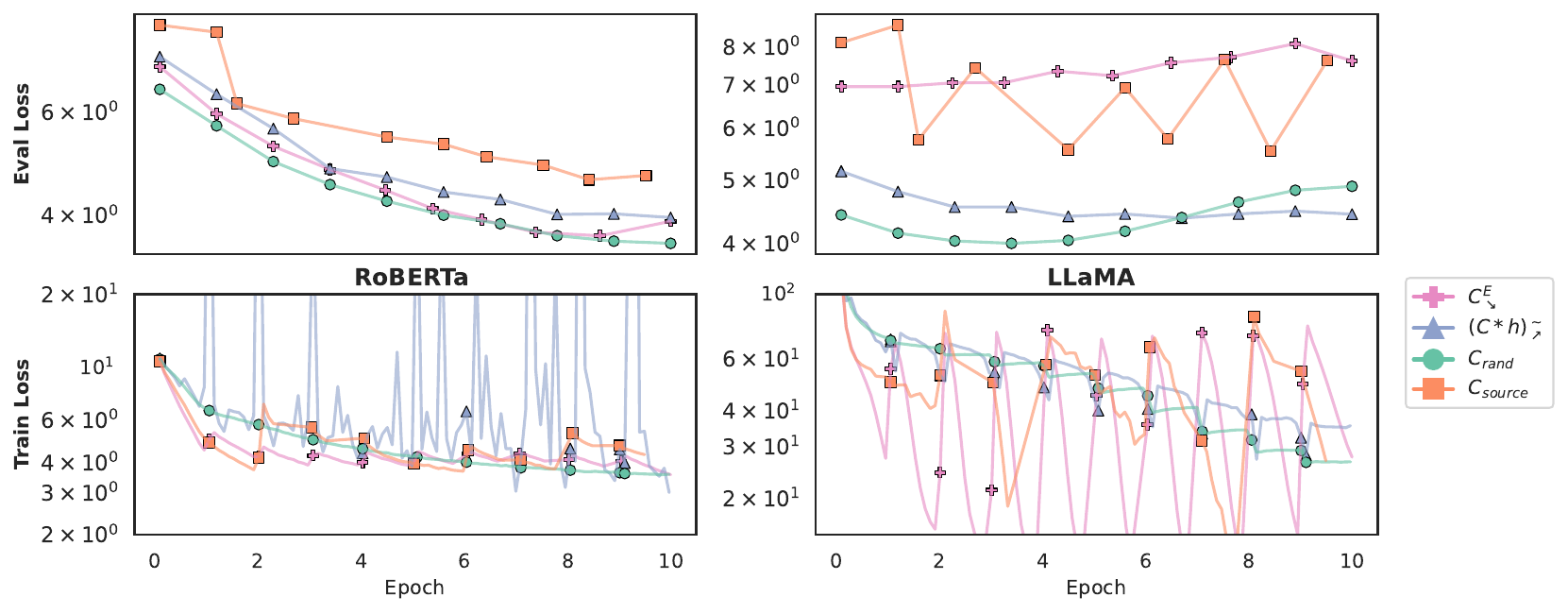}
  \caption{Train- and evaluation loss of baselines and influence curricula for $D_{stratified}$.}
  \label{train-eval-loss-aio}
\end{figure*}
\subsection{Loss Trajectories}
We provide training- and evaluation loss trajectories for a subset of our models in Figure \ref{train-eval-loss-aio}, and the remaining ones in Appendix \ref{appendix-benchmark-results-loss-trajectories}. For one RoBERTa model ($C^E_{\nearrow}$ on $D_{2024}$) and 9 Llama models ($D_{2024}$: \{$C^E_{\nearrow}, C_A, C_{source}, C_{MATTR}, C_{rand}$\}, $D_{stratified}$: \{$C^E_{\nearrow}, C^E_{\searrow}, C^{\{50\}}_{\nearrow}, C_{rand}\}$) we measure higher evaluation loss at the end of training compared to the beginning, suggesting training divergence.

We observe \textbf{substantial training loss spikes}, which in non-curriculum learning often indicates training instability \cite{li_stability-efficiency_2022}.
However, as evident in Figure \ref{train-eval-loss-aio}, the model that performs best in benchmarks ($(C*h)^\sim_\nearrow$, RoBERTa) exhibits more severe training loss-spikes than the worse performing $C_{source}$, $C^E_{\searrow}$ or $C_{random}$.
We extend this analysis to all 84 models, calculating the Spearman rank correlation between a curriculum's gain in benchmark performance (over training in random order) and the \textit{loss-ratio} (a measure of training instability; \citealp{li_stability-efficiency_2022}) in Appendix \ref{appendix-loss-trajectories-experiment}. We find no significant negative rank correlation for any dataset\footnote{$D_{2024}$:  0.177, $D_{equitoken}$: 0.096, $D_{stratified}$: 0.197}, indicating that at least within the limited number of epochs we train for, training loss trajectories appear \textbf{less informative of downstream performance} compared to training in random order.
\subsection{Document Order}
\begin{figure}[t]
  \includegraphics[width=0.9\columnwidth, trim=0cm 0.3cm 0cm 0.5cm, clip]{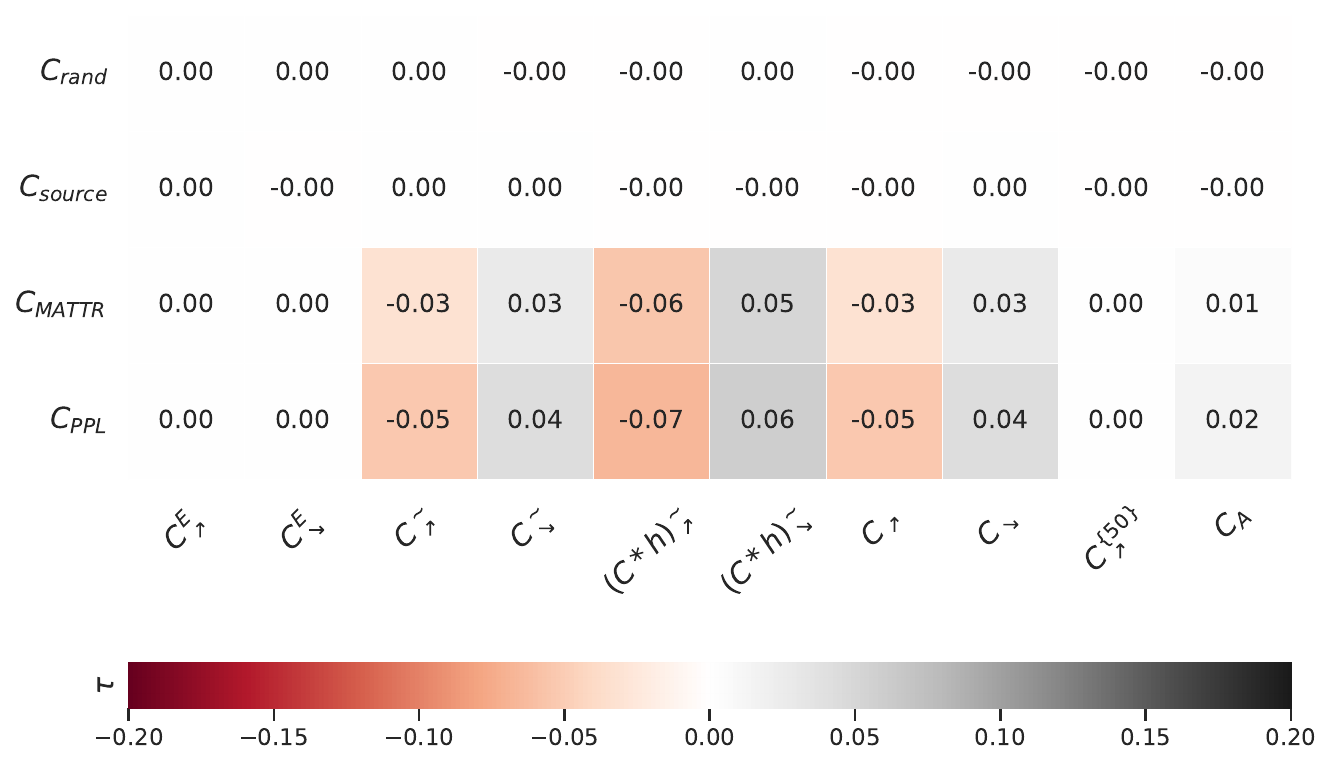}
  \caption{Rank-similarity between influence-curricula and baselines: Mean Kendall $\tau_b$.}
  \label{influence-plot}
\end{figure}
We additionally explore how the ordering of examples under influence curricula correlates with ordering of existing heuristics.
We use Kendall's $\tau$, calculated on a per-epoch basis as documents are shown multiple times during training.\footnote{As documents may also be visited multiple times within an epoch, we use tau-b \cite{kendall_treatment_1945} to account for ties. We truncate the longer of the two curricula where necessary.}
Curricula sorted by decreasing influence ($C^{\sim}_{\searrow}$, $C_{\searrow}$, $(C*h)^{\sim}_{\searrow}$) show significantly stronger correlations with both $C_{MATTR}$ and $C_{PPL}$ than curricula sorted by increasing influence ($C_{MATTR}:+0.047^*$, $C_{PPL}:+0.084^*$). This suggests that our influence measure \textbf{may be inversely related to example difficulty} as defined by these curricula (i.e., higher influence implies lower difficulty). 
Rank correlation between any type of influence curriculum and $C_{rand}$, as well as between influence curricula and $C_{source}$ is negligible, which is to be expected as we shuffle these within epochs or stages respectively.
Convolving with a log-norm filter before sorting ($(C*h)^\sim_\searrow$) has a marginal positive, but insignificant effect on the similarity to baselines (+0.016 w.r.t. $C^\sim_\searrow$ for $C_{MATTR}$, +0.013 w.r.t $C_{PPL}$).
\section{Discussion}\label{discussion}
Our results indicate that curricula based on training data influence estimates can be viable from a performance perspective; however, they are only so if paired with non-developmentally plausible coverage strategies (i.e., ones roughly inspired by how humans acquire language), in which the full training data is visited once per epoch: 
When specifically comparing the handcrafted- ($C_{source}$), and the two synthetic source-difficulty curricula ($C^{E}_{\nearrow}$, $C^{E}_{\searrow}$), it is evident that our sorting strategy based on training dynamics was unable to compensate for this less effective human-centered form of scheduling in terms of performance. Future work should therefore explore coverage strategies that more effectively balance model performance and developmentally plausible scheduling.

The observation that the ascending versions of the same strategy do not consistently outperform the descending versions (e.g., $C_\searrow$ and $C_\nearrow$) and vice versa suggests that the observed increase in performance might not stem from the specific sorting order (by increasing or decreasing influence), but rather from an \textbf{improved grouping of examples:} examples of similar influence are more likely located in the same batch. 
This would also explain the competitive performance of sorting by the model-agnostic difficulty heuristic $C_{MATTR}$ on $D_{2024}$ and $D_{stratified}$. 
\section{Conclusion}
In this work, we study curriculum learning for language model pretraining and propose a novel type of curricula based on training data influence, which \textbf{outperforms training in random order} by up to 12.42 pp for RoBERTa models ($C_{\{50\}}$, $D_{2024}$) and up to 4.62 pp for Llama models ($C_{\nearrow}$, $D_{stratified}$). 
In contrast to recent experiments with handcrafted curricula, our results indicate that curriculum learning with our method has \textbf{potential to improve data efficiency in low-resource settings}.

Through an analysis of the data distribution in our curricula derived from influence estimates, we find that \textbf{their source composition does not strongly vary over time}, contrasting that of existing source-difficulty curricula, which are typically designed to decrease the proportion of child-directed speech in later epochs (replacing it with more complex text).
Furthermore, by conducting an analysis of training- and evaluation loss trajectories, we have observed that the severe spikes in training loss seen with this form of curriculum learning are not significantly correlated with model performance on downstream benchmarks. Lastly, we explore how the ordering of examples with influence curricula correlates with existing sorting heuristics, finding that our measure is \textbf{inversely correlated to example difficulty} (i.e., higher influence implies lower difficulty).
In conclusion, our results suggest that curricula based on training data influence estimates can be viable from a performance perspective, but, their success may be attributed to training dynamics rather than increased developmental plausibility.
\section*{Limitations}\label{limitations}
We use a two-step approach to estimating training data influence: we first pre-train a model in random order,  and subsequently extract the loss-gradients we utilize for influence estimation (one example at a time). We opted for this implementation to simplify our experimental setup, as our primary focus was on studying curriculum learning rather than minimizing training time. 
To improve computational efficiency within our framework, one could reuse (mini-batch) gradients from model training for influence estimation. We provide additional details on runtime in Appendix \ref{appendix-implementation-details}.

In Section \ref{datamixanalysis}, where we study the data mix of our curricula, we observe that our influence curricula are highly sensitive to the source distribution of the dataset.
If future work has an intention to use a similar influence estimation method for data cleaning or selection (as we did in $C^{\{50\}}$), it should explore measures to ensure appropriate data balancing. In our setup, the failure to do so primarily results in lower benchmark performance for $C^{\{50\}}$.

Lastly, our experiments are based on relatively small language models and datasets due to the lack of large-scale pre-training datasets that both cover and categorize examples across different difficulty levels. However, with $D_{2024}$ we include a dataset that is widely used and studied through the BabyLM challenge (see \citealp{charpentier_babylm_2025}).
\section*{Acknowledgments}
The present research was funded by the Go!Digital 3.0 grant program of the Austrian Academy of Sciences (GD3.0\_2021-18\_CogML) and by the Vienna Science and Technology Fund (WWTF)[10.47379/VRG19008] "Knowledge-infused Deep Learning for Natural Language Processing".
\bibliography{custom}
\clearpage
\appendix
\section{Implementation Details}\label{appendix-implementation-details}
\paragraph{Influence Estimation}\mbox{}\\
To enable influence estimation for the RoBERTa models, which are trained with dynamic masking (tokens are masked differently at each epoch), we implement a custom \textit{Data Collator} for use with the Hugging Face \textit{Trainer}: This collator makes masking reproducible by computing a hash based on the document and the epoch number. 

\paragraph{Runtime}\label{appendix-runtime}\mbox{}\\
Pre-training of all 84 models took 195 hours on 4 NVIDIA H100 GPUs (approximately 2h20m per model).
The runtime of the influence estimation step, which is only required once per dataset, depends on the number of documents.
On average (across model architectures and datasets) it amounts to roughly 44.3h if estimation is run sequentially for each checkpoint, or just under 5h if run in parallel.
Sequential runtime would amount to 7h45h for $D_{equitoken}$, 109h for $D_{2024}$ (both ran on NVIDIA H100 GPUs), and 149h30min on $D_{stratified}$ (ran on a lower-spec NVIDIA V100 GPUs), totaling 266 GPU hours overall.

\begin{table}[ht]
\footnotesize
\centering
\begin{tabular}{|l|c|c|}
\hline
 & \textbf{RoBERTa} & \textbf{LLaMA} \\
\hline
Vocabulary size & \multicolumn{2}{c|}{52k} \\
Hidden size & \multicolumn{2}{c|}{768} \\
Number of layers & \multicolumn{2}{c|}{12} \\
Number of attention heads & \multicolumn{2}{c|}{12} \\
Initializer range & \multicolumn{2}{c|}{0.02} \\
Tie word embeddings & \multicolumn{2}{c|}{True} \\
\hline
Max position embeddings & 514 & \texttt{256} \\
Intermediate (FFN) size & 3072 & 2048 \\
Norm epsilon & 1e-5 & 1e-6 \\
Attention dropout & 0.1 & 0 \\
Activation function & \texttt{gelu} & \texttt{silu} \\
Hidden dropout & 0.1 & - \\
\hline
FP16 & \multicolumn{2}{c|}{False}\\
Per Device Batch Size & \multicolumn{2}{c|}{32}\\
Gradient Accumulation Steps & \multicolumn{2}{c|}{16} \\
GPUs & \multicolumn{2}{c|}{4} \\
Adam $\beta_1$ & \multicolumn{2}{c|}{0.9}\\
Adam $\beta_2$ & \multicolumn{2}{c|}{0.98}\\
Adam $\epsilon$ & \multicolumn{2}{c|}{1e-6}\\
Weight Decay $\epsilon$ & \multicolumn{2}{c|}{0.01}\\

\hline
Learning rate & 5e-4 & 7e-4 \\
Scheduler & polynomial & cosine \\
\hline
\end{tabular}
\caption{Training parameters used for all models.}
\label{training-params}
\end{table}

\section{Loss Trajectories}\label{appendix-loss-trajectories-experiment}
Our curricula sort examples based on their influence, which may inadvertently reduce example diversity within training batches. We hypothesize that this led to the substantial training loss spikes observed. 
While one can measure loss during training with a separate evaluation set (as we have done), this adds significant overhead during training.
To analyze whether training loss spikes are still indicative of training instability for curriculum learning, i.e. wether their severity ultimately impacts benchmark performance, we employ the \textit{loss ratio} metric proposed by \citet{li_stability-efficiency_2022}, as a measure of training instability, which compares the loss at the current step $s$ to the lowest loss achieved in any prior step: $lr(s) = \frac{\ell(s)}{\min_{s' < s} \ell(s')}$. Intuitively (if training in random order), one would expect models with high loss ratios to have lower benchmark performance. However, an analysis of the corelation between a curriculum's gain in benchmark performance (over training in random order) and this loss-ratio indeed does not reveal a significant negative Spearman rank correlation for any dataset: $D_{2024}$:  0.177; $D_{equitoken}$: 0.096; $D_{stratified}$: 0.197. 

\FloatBarrier

\onecolumn
\section{Complementary Figures}\label{appendix-complementary-figures}
This section presents complementary figures for RoBERTa or Llama models, with the respective other model type included in the main body of our paper.
\begin{figure}[ht]
\centering
  \includegraphics[width=0.5\textwidth]{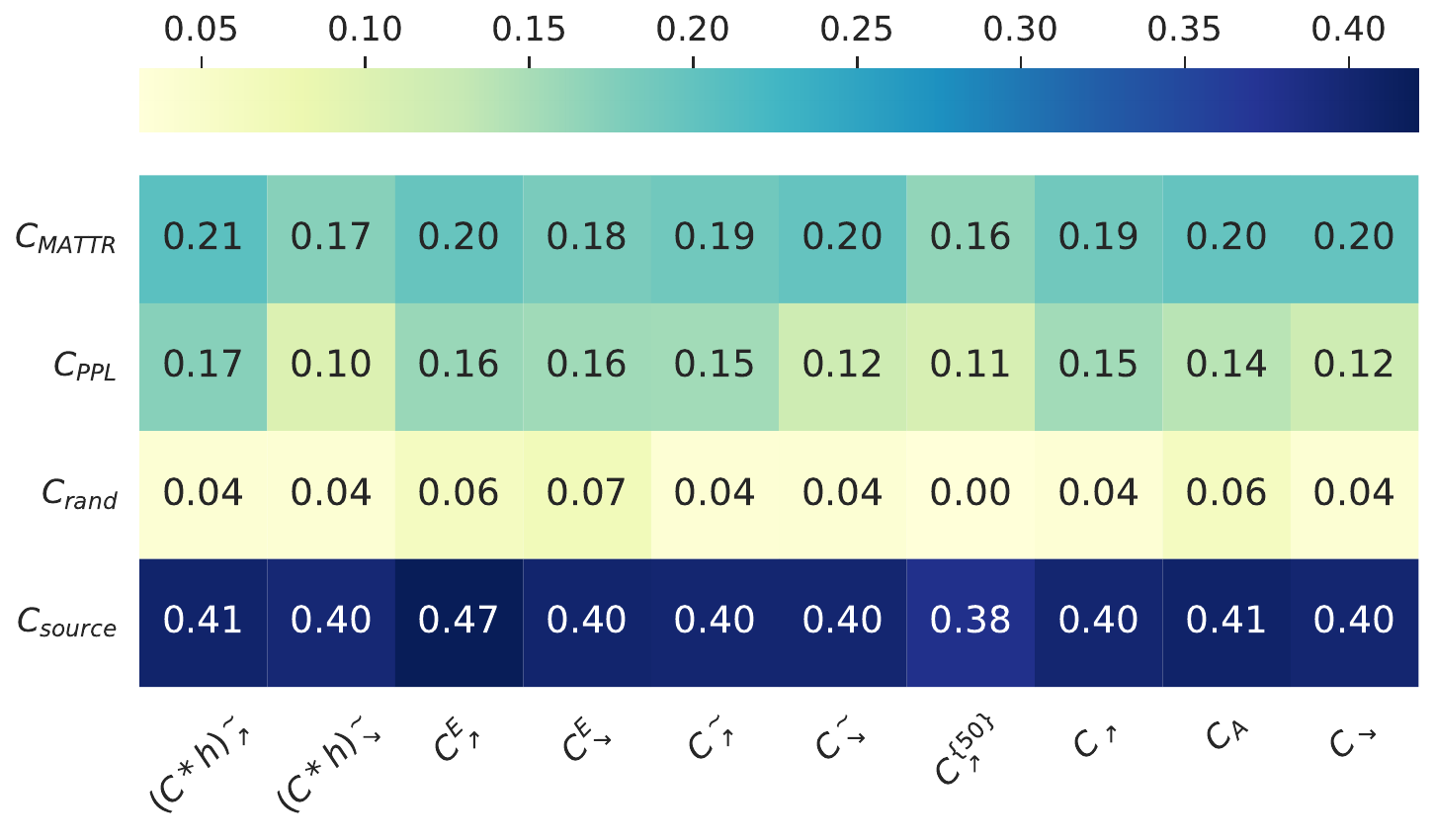}
  \caption{Comparison of curriculum stage distributions: Average Jensen–Shannon divergence between 1000 segments of two given curricula for RoBERTa models. Lower values indicate more similar stage distributions.}
  \label{jensen-shannon-plot-roberta}
\end{figure}
\begin{figure}[ht]
  \includegraphics[width=\textwidth]{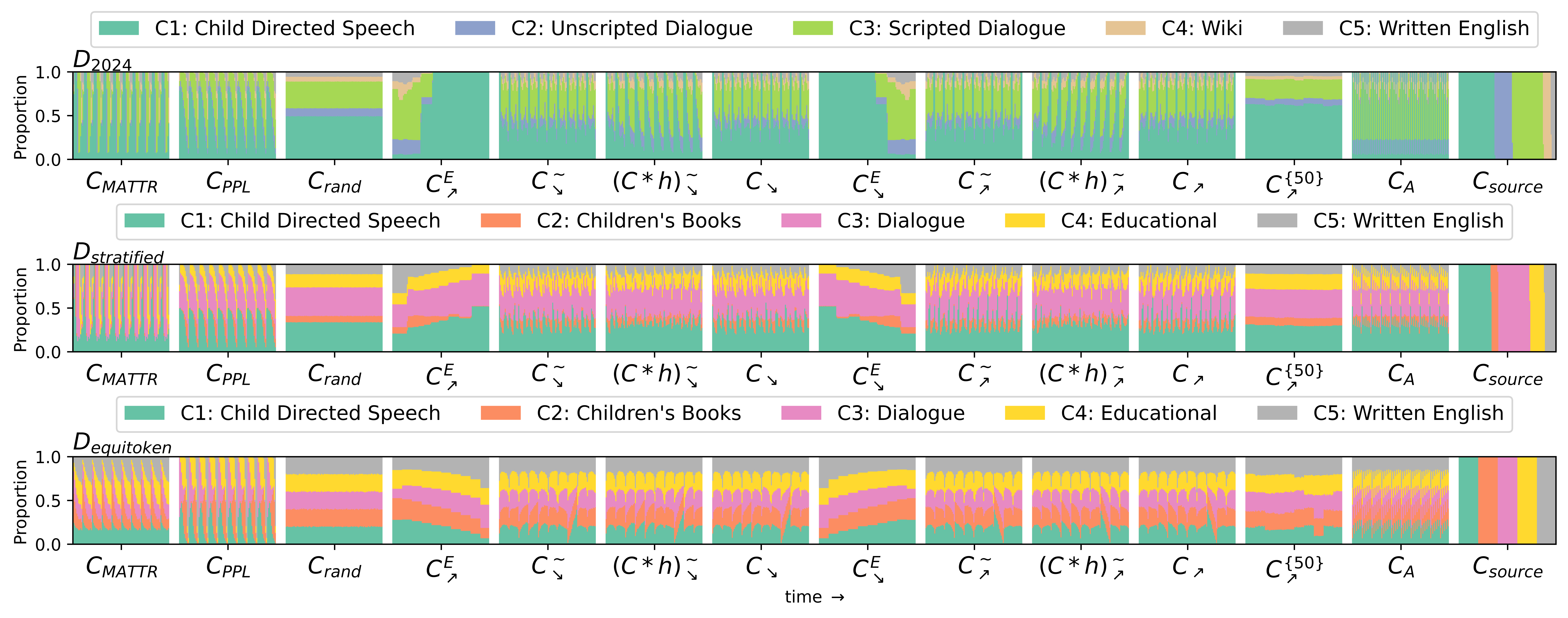}
  \caption{Dataset mix of curricula for RoBERTa models. We trace back documents to the stages defined in Table \ref{tab:source-difficulty-combined}. }
  \label{influence-plosst-roberta}
\end{figure}
\FloatBarrier

\section{Full Benchmark Results and Loss Trajectories}\label{appendix-benchmark-results-loss-trajectories}
    \begin{figure*}[t]
  \centering
  \begin{minipage}{0.86\textwidth}
    \centering
    
    \includegraphics[width=\textwidth]{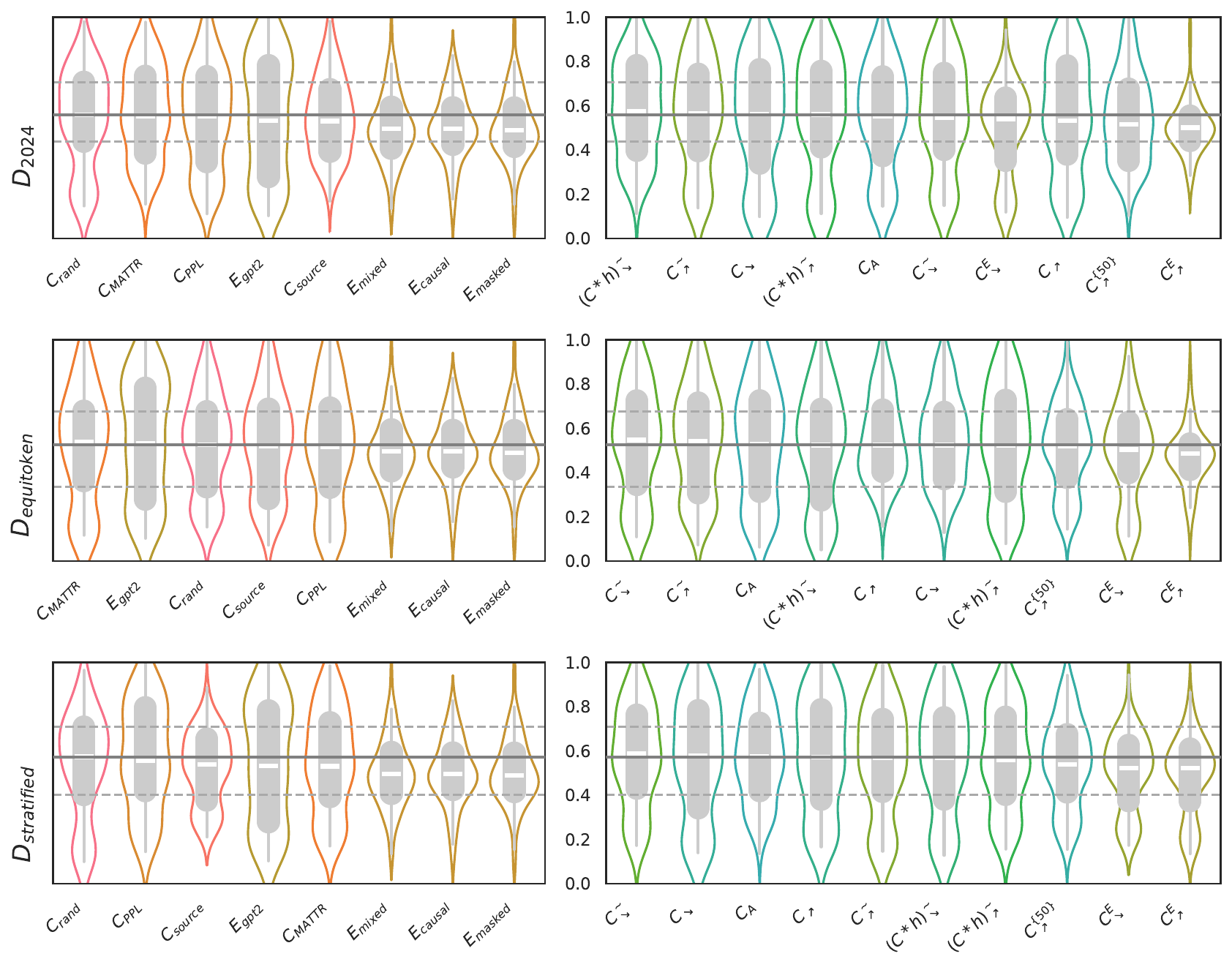}
    \caption{Benchmark results for Llama models.} 
    \label{violinplot_llama}
  \end{minipage}
  \\[0.0cm] 
  \begin{minipage}{0.86\textwidth}
    \centering
    
    \includegraphics[width=\textwidth]{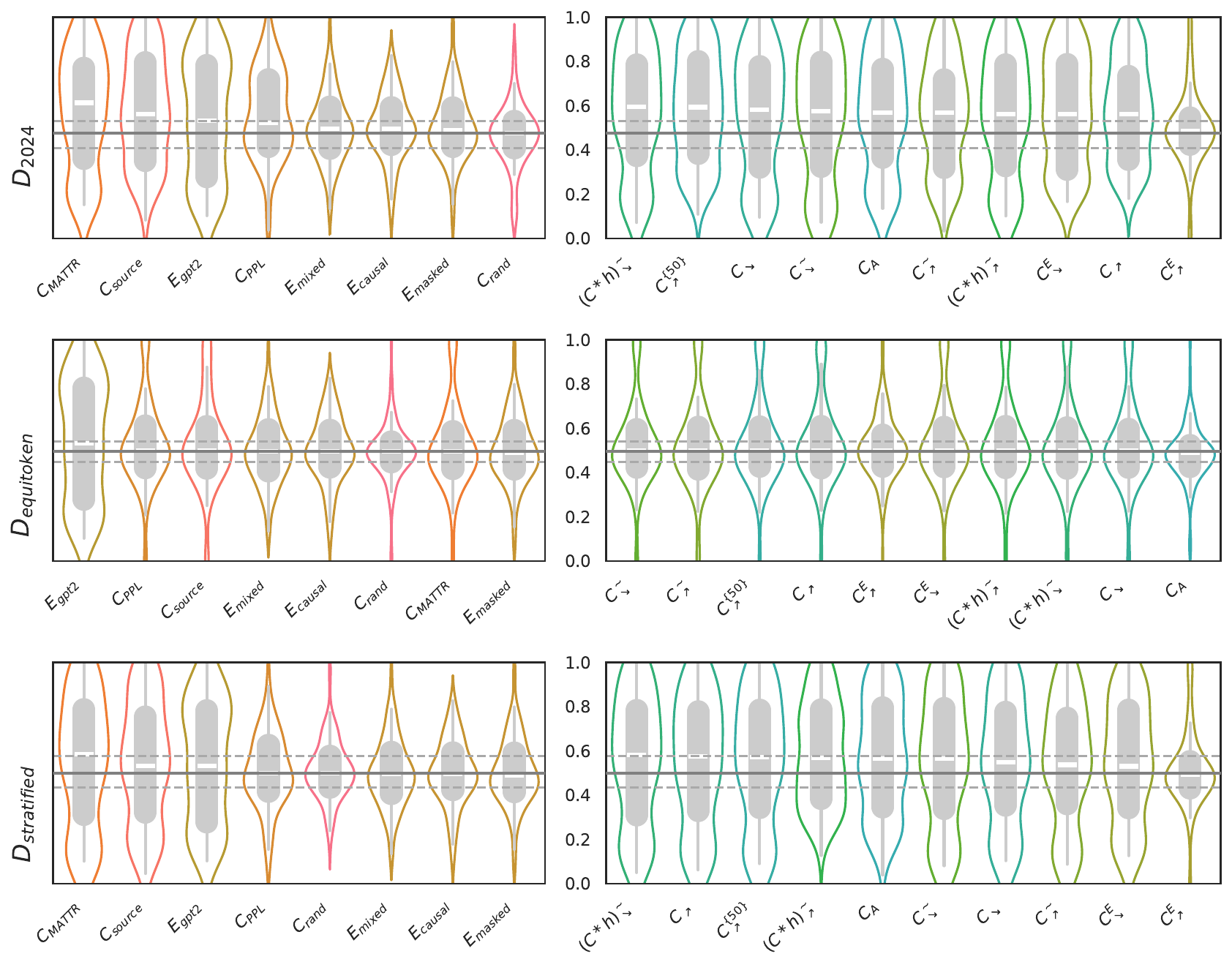}
    \caption{Benchmark results for RoBERTa models.}
    \label{violinplot_roberta}
  \end{minipage}
\end{figure*}
\footnotesize{
\begin{longtable}{lllllrl}
\caption{Macro-average gain in accuracy over the corresponding random curriculum.} \\
\toprule
Curriculum & Dataset & Architecture & Improvement & p-val & Model acc & Random acc \\
\midrule
\endfirsthead
\caption*{Continued from previous page} \\
\toprule
Curriculum & Dataset & Architecture & Improvement & p-val & Model acc & Random acc \\
\midrule
\endhead
\midrule
\multicolumn{7}{r}{Continued on next page} \\
\endfoot
\bottomrule
\endlastfoot
$C_{rand}$ & $D_{2024}$ & RoBERTa & +0.00 pp & - & 0.466 & - \\
$C_{rand}$ & $D_{equitoken}$ & RoBERTa & +0.00 pp & - & 0.492 & - \\
$C_{rand}$ & $D_{stratified}$ & RoBERTa & +0.00 pp & - & 0.512 & - \\
$C_{rand}$ & $D_{equitoken}$ & Llama & +0.00 pp & - & 0.523 & - \\
$C_{rand}$ & $D_{stratified}$ & Llama & +0.00 pp & - & 0.536 & - \\
$C_{rand}$ & $D_{2024}$ & Llama & +0.00 pp & - & 0.541 & - \\
\hline \hline
$C^E_{\nearrow}$ & $D_{equitoken}$ & Llama & -5.02 pp** & 0.033 & 0.473 & 0.523 \\
$C^E_{\searrow}$ & $D_{2024}$ & Llama & -4.84 pp*** & 0.004 & 0.493 & 0.541 \\
$C^E_{\nearrow}$ & $D_{stratified}$ & Llama & -4.79 pp*** & 0.005 & 0.488 & 0.536 \\
$C^E_{\nearrow}$ & $D_{2024}$ & Llama & -3.83 pp* & 0.065 & 0.503 & 0.541 \\
$C^E_{\searrow}$ & $D_{stratified}$ & Llama & -3.11 pp*** & 0.002 & 0.504 & 0.536 \\
$C^E_{\searrow}$ & $D_{equitoken}$ & Llama & -3.10 pp & 0.100 & 0.492 & 0.523 \\
$(C*h)^{\sim}_{\searrow}$ & $D_{equitoken}$ & Llama & -1.82 pp & 0.400 & 0.505 & 0.523 \\
$C_{source}$ & $D_{stratified}$ & Llama & -1.39 pp & 0.167 & 0.522 & 0.536 \\
$C^{\{50\}}_{\nearrow}$ & $D_{equitoken}$ & Llama & -1.24 pp & 0.504 & 0.511 & 0.523 \\
$C_{MATTR}$ & $D_{equitoken}$ & Llama & -0.72 pp & 0.293 & 0.516 & 0.523 \\
$C_{PPL}$ & $D_{equitoken}$ & Llama & -0.65 pp & 0.431 & 0.517 & 0.523 \\
$C_{A}$ & $D_{equitoken}$ & RoBERTa & -0.55 pp & 0.726 & 0.487 & 0.492 \\
$C_{PPL}$ & $D_{stratified}$ & RoBERTa & -0.28 pp & 0.877 & 0.510 & 0.512 \\
$C_{source}$ & $D_{equitoken}$ & Llama & -0.12 pp & 0.856 & 0.522 & 0.523 \\
$C_{\searrow}$ & $D_{2024}$ & Llama & -0.02 pp & 0.991 & 0.541 & 0.541 \\
$C_{A}$ & $D_{2024}$ & Llama & +0.17 pp & 0.796 & 0.543 & 0.541 \\
$C^{\{50\}}_{\nearrow}$ & $D_{2024}$ & Llama & +0.21 pp & 0.918 & 0.543 & 0.541 \\
$C^E_{\nearrow}$ & $D_{stratified}$ & RoBERTa & +0.36 pp & 0.801 & 0.516 & 0.512 \\
$(C*h)^{\sim}_{\nearrow}$ & $D_{equitoken}$ & Llama & +0.42 pp & 0.848 & 0.527 & 0.523 \\
$C_{A}$ & $D_{equitoken}$ & Llama & +0.53 pp & 0.813 & 0.528 & 0.523 \\
$C_{PPL}$ & $D_{2024}$ & Llama & +0.72 pp & 0.317 & 0.548 & 0.541 \\
$C^{\{50\}}_{\nearrow}$ & $D_{stratified}$ & Llama & +0.92 pp & 0.619 & 0.545 & 0.536 \\
$C_{source}$ & $D_{2024}$ & Llama & +1.07 pp & 0.242 & 0.552 & 0.541 \\
$(C*h)^{\sim}_{\nearrow}$ & $D_{2024}$ & Llama & +1.29 pp & 0.504 & 0.554 & 0.541 \\
$C^{\sim}_{\searrow}$ & $D_{equitoken}$ & Llama & +1.31 pp & 0.150 & 0.536 & 0.523 \\
$C^{\sim}_{\searrow}$ & $D_{2024}$ & Llama & +1.37 pp & 0.477 & 0.555 & 0.541 \\
$C^{\sim}_{\nearrow}$ & $D_{equitoken}$ & Llama & +1.50 pp & 0.494 & 0.538 & 0.523 \\
$C^{\sim}_{\nearrow}$ & $D_{stratified}$ & Llama & +1.73 pp*** & 0.007 & 0.553 & 0.536 \\
$C^{\sim}_{\nearrow}$ & $D_{2024}$ & Llama & +1.77 pp & 0.362 & 0.559 & 0.541 \\
$C_\nearrow$ & $D_{2024}$ & Llama & +1.78 pp & 0.371 & 0.559 & 0.541 \\
$C_{MATTR}$ & $D_{stratified}$ & Llama & +1.86 pp** & 0.029 & 0.554 & 0.536 \\
$C^E_{\nearrow}$ & $D_{equitoken}$ & RoBERTa & +1.93 pp & 0.236 & 0.512 & 0.492 \\
$C_{\searrow}$ & $D_{equitoken}$ & Llama & +2.29 pp*** & 0.006 & 0.546 & 0.523 \\
$C_{\searrow}$ & $D_{stratified}$ & Llama & +2.37 pp*** & 0.002 & 0.559 & 0.536 \\
$(C*h)^{\sim}_{\searrow}$ & $D_{stratified}$ & Llama & +2.41 pp*** & 0.001 & 0.560 & 0.536 \\
$C_{MATTR}$ & $D_{equitoken}$ & RoBERTa & +2.62 pp & 0.138 & 0.518 & 0.492 \\
$C^E_{\nearrow}$ & $D_{2024}$ & RoBERTa & +3.02 pp & 0.124 & 0.496 & 0.466 \\
$C_{MATTR}$ & $D_{2024}$ & Llama & +3.07 pp*** & 0.000 & 0.572 & 0.541 \\
$(C*h)^{\sim}_{\nearrow}$ & $D_{stratified}$ & Llama & +3.08 pp & 0.122 & 0.566 & 0.536 \\
$(C*h)^{\sim}_{\searrow}$ & $D_{equitoken}$ & RoBERTa & +3.10 pp & 0.123 & 0.523 & 0.492 \\
$C^{\sim}_{\searrow}$ & $D_{equitoken}$ & RoBERTa & +3.12 pp* & 0.079 & 0.523 & 0.492 \\
$C_\nearrow$ & $D_{equitoken}$ & Llama & +3.16 pp & 0.142 & 0.555 & 0.523 \\
$C^{\sim}_{\nearrow}$ & $D_{equitoken}$ & RoBERTa & +3.31 pp* & 0.077 & 0.525 & 0.492 \\
$C^{\sim}_{\nearrow}$ & $D_{stratified}$ & RoBERTa & +3.32 pp & 0.166 & 0.546 & 0.512 \\
$C_\nearrow$ & $D_{stratified}$ & RoBERTa & +3.51 pp & 0.140 & 0.548 & 0.512 \\
$C_{\searrow}$ & $D_{equitoken}$ & RoBERTa & +3.57 pp** & 0.050 & 0.528 & 0.492 \\
$C_{MATTR}$ & $D_{stratified}$ & RoBERTa & +3.81 pp & 0.126 & 0.551 & 0.512 \\
$C_{source}$ & $D_{stratified}$ & RoBERTa & +3.89 pp & 0.120 & 0.551 & 0.512 \\
$C_{PPL}$ & $D_{equitoken}$ & RoBERTa & +3.92 pp** & 0.032 & 0.531 & 0.492 \\
$C_{PPL}$ & $D_{stratified}$ & Llama & +3.97 pp*** & 0.000 & 0.575 & 0.536 \\
$(C*h)^{\sim}_{\nearrow}$ & $D_{equitoken}$ & RoBERTa & +4.00 pp* & 0.050 & 0.532 & 0.492 \\
$C_{source}$ & $D_{equitoken}$ & RoBERTa & +4.12 pp* & 0.052 & 0.533 & 0.492 \\
$C^E_{\searrow}$ & $D_{equitoken}$ & RoBERTa & +4.16 pp** & 0.041 & 0.534 & 0.492 \\
$C_{\searrow}$ & $D_{stratified}$ & RoBERTa & +4.16 pp* & 0.079 & 0.554 & 0.512 \\
$C_{A}$ & $D_{stratified}$ & Llama & +4.18 pp*** & 0.000 & 0.577 & 0.536 \\
$C^{\sim}_{\searrow}$ & $D_{stratified}$ & Llama & +4.18 pp*** & 0.000 & 0.577 & 0.536 \\
$C^{\sim}_{\searrow}$ & $D_{stratified}$ & RoBERTa & +4.26 pp* & 0.094 & 0.555 & 0.512 \\
$(C*h)^{\sim}_{\searrow}$ & $D_{2024}$ & Llama & +4.34 pp** & 0.028 & 0.584 & 0.541 \\
$C^{\{50\}}_{\nearrow}$ & $D_{equitoken}$ & RoBERTa & +4.36 pp** & 0.039 & 0.536 & 0.492 \\
$C^E_{\searrow}$ & $D_{stratified}$ & RoBERTa & +4.40 pp* & 0.052 & 0.556 & 0.512 \\
$(C*h)^{\sim}_{\searrow}$ & $D_{stratified}$ & RoBERTa & +4.47 pp* & 0.072 & 0.557 & 0.512 \\
$C^{\{50\}}_{\nearrow}$ & $D_{stratified}$ & RoBERTa & +4.52 pp* & 0.067 & 0.558 & 0.512 \\
$C_\nearrow$ & $D_{equitoken}$ & RoBERTa & +4.54 pp** & 0.031 & 0.538 & 0.492 \\
$C_\nearrow$ & $D_{stratified}$ & Llama & +4.62 pp*** & 0.000 & 0.582 & 0.536 \\
$C_{A}$ & $D_{stratified}$ & RoBERTa & +6.67 pp*** & 0.004 & 0.579 & 0.512 \\
$(C*h)^{\sim}_{\nearrow}$ & $D_{stratified}$ & RoBERTa & +7.96 pp*** & 0.000 & 0.592 & 0.512 \\
$C_{\searrow}$ & $D_{2024}$ & RoBERTa & +8.72 pp*** & 0.004 & 0.553 & 0.466 \\
$(C*h)^{\sim}_{\nearrow}$ & $D_{2024}$ & RoBERTa & +8.74 pp*** & 0.002 & 0.553 & 0.466 \\
$C^{\sim}_{\searrow}$ & $D_{2024}$ & RoBERTa & +9.13 pp*** & 0.002 & 0.557 & 0.466 \\
$C^{\sim}_{\nearrow}$ & $D_{2024}$ & RoBERTa & +9.36 pp*** & 0.000 & 0.559 & 0.466 \\
$C_{PPL}$ & $D_{2024}$ & RoBERTa & +9.49 pp*** & 0.000 & 0.561 & 0.466 \\
$C_{A}$ & $D_{2024}$ & RoBERTa & +10.19 pp*** & 0.001 & 0.568 & 0.466 \\
$(C*h)^{\sim}_{\searrow}$ & $D_{2024}$ & RoBERTa & +10.71 pp*** & 0.000 & 0.573 & 0.466 \\
$C_{MATTR}$ & $D_{2024}$ & RoBERTa & +10.97 pp*** & 0.000 & 0.575 & 0.466 \\
$C^E_{\searrow}$ & $D_{2024}$ & RoBERTa & +10.98 pp*** & 0.000 & 0.576 & 0.466 \\
$C_\nearrow$ & $D_{2024}$ & RoBERTa & +11.00 pp*** & 0.000 & 0.576 & 0.466 \\
$C_{source}$ & $D_{2024}$ & RoBERTa & +11.77 pp*** & 0.000 & 0.583 & 0.466 \\
$C^{\{50\}}_{\nearrow}$ & $D_{2024}$ & RoBERTa & +12.42 pp*** & 0.000 & 0.590 & 0.466 \\
$E_{mixed}$ & ext & gpt-bert & - & - & 0.498 & - \\
$E_{causal}$ & ext & gpt-bert & - & - & 0.502 & - \\
$E_{masked}$ & ext & gpt-bert & - & - & 0.504 & - \\
$E_{gpt2}$ & ext & - & - & - & 0.551 & - \\
\end{longtable}

  }
\begin{landscape}
\footnotesize{
\begin{longtable}{lllllllll}
\caption{Macro accuracy for Llama models across tasks, per benchmark and overall. $E_\cdot$ denotes baseline models from the BabyLM challange, the fine-tuning evaluation pipeline fails for the $E_{gpt2}$ model.} \\
\toprule
Curriculum & Dataset & (Super) GLUE & blimp\_filtered & supplement\_filtered & entity\_tracking & ewok\_filtered & wug\_adj\_nominalization & Macro acc \\
\midrule
\endfirsthead
\caption*{Continued from previous page} \\
\toprule
Curriculum & Dataset & (Super) GLUE & blimp\_filtered & supplement\_filtered & entity\_tracking & ewok\_filtered & wug\_adj\_nominalization & Macro acc \\
\midrule
\endhead
\midrule
\multicolumn{9}{r}{Continued on next page} \\
\endfoot
\bottomrule
\endlastfoot
$(C*h)^{\sim}_{\searrow}$ & $D_{2024}$ & 0.579 & 0.688 & 0.559 & 0.302 & 0.509 & 0.570 & \textbf{0.584} \\
$C_\nearrow$ & $D_{stratified}$ & 0.573 & \textbf{0.715} & 0.546 & 0.208 & 0.503 & 0.600 & 0.582 \\
$C_{A}$ & $D_{stratified}$ & 0.575 & 0.675 & 0.575 & 0.306 & 0.507 & 0.560 & 0.577 \\
$C^{\sim}_{\searrow}$ & $D_{stratified}$ & 0.573 & 0.695 & 0.558 & 0.242 & \textbf{0.519} & 0.495 & 0.577 \\
$C_{PPL}$ & $D_{stratified}$ & 0.573 & 0.696 & 0.532 & 0.239 & 0.510 & 0.480 & 0.575 \\
$C_{MATTR}$ & $D_{2024}$ & 0.573 & 0.671 & 0.551 & 0.295 & 0.507 & 0.550 & 0.572 \\
$(C*h)^{\sim}_{\nearrow}$ & $D_{stratified}$ & 0.572 & 0.678 & 0.567 & 0.245 & 0.501 & 0.540 & 0.566 \\
$(C*h)^{\sim}_{\searrow}$ & $D_{stratified}$ & 0.573 & 0.691 & 0.542 & 0.169 & 0.507 & 0.510 & 0.560 \\
$C_{\searrow}$ & $D_{stratified}$ & 0.567 & 0.694 & 0.533 & 0.164 & 0.512 & 0.420 & 0.559 \\
$C_\nearrow$ & $D_{2024}$ & 0.575 & 0.686 & 0.566 & 0.184 & 0.494 & 0.500 & 0.559 \\
$C^{\sim}_{\nearrow}$ & $D_{2024}$ & 0.571 & 0.683 & 0.566 & 0.184 & 0.506 & 0.565 & 0.559 \\
$C^{\sim}_{\searrow}$ & $D_{2024}$ & 0.571 & 0.679 & 0.571 & 0.176 & 0.506 & 0.500 & 0.555 \\
$C_\nearrow$ & $D_{equitoken}$ & 0.575 & 0.618 & 0.514 & 0.389 & 0.489 & 0.555 & 0.555 \\
$C_{MATTR}$ & $D_{stratified}$ & 0.571 & 0.663 & 0.539 & 0.227 & 0.516 & 0.495 & 0.554 \\
$(C*h)^{\sim}_{\nearrow}$ & $D_{2024}$ & 0.570 & 0.692 & 0.536 & 0.136 & 0.511 & 0.515 & 0.554 \\
$C^{\sim}_{\nearrow}$ & $D_{stratified}$ & 0.568 & 0.684 & 0.514 & 0.169 & 0.499 & 0.535 & 0.553 \\
$C_{source}$ & $D_{2024}$ & 0.576 & 0.628 & 0.503 & 0.336 & 0.505 & 0.560 & 0.552 \\
$E_{gpt2}$ & ext & nan & 0.673 & \textbf{0.591} & 0.189 & 0.498 & 0.390 & 0.551 \\
$C_{PPL}$ & $D_{2024}$ & 0.582 & 0.655 & 0.508 & 0.226 & 0.499 & 0.655 & 0.548 \\
$C_{\searrow}$ & $D_{equitoken}$ & 0.577 & 0.615 & 0.528 & 0.336 & 0.501 & 0.685 & 0.546 \\
$C^{\{50\}}_{\nearrow}$ & $D_{stratified}$ & 0.575 & 0.633 & 0.540 & 0.267 & 0.510 & 0.635 & 0.545 \\
$C_{A}$ & $D_{2024}$ & 0.573 & 0.660 & 0.520 & 0.178 & 0.506 & 0.635 & 0.543 \\
$C^{\{50\}}_{\nearrow}$ & $D_{2024}$ & 0.572 & 0.618 & 0.541 & 0.314 & 0.497 & 0.635 & 0.543 \\
$C_{rand}$ & $D_{2024}$ & 0.572 & 0.658 & 0.497 & 0.193 & 0.500 & 0.440 & 0.541 \\
$C_{\searrow}$ & $D_{2024}$ & 0.573 & 0.674 & 0.521 & 0.133 & 0.500 & 0.530 & 0.541 \\
$C^{\sim}_{\nearrow}$ & $D_{equitoken}$ & 0.577 & 0.634 & 0.561 & 0.234 & 0.503 & 0.465 & 0.538 \\
$C_{rand}$ & $D_{stratified}$ & 0.576 & 0.662 & 0.517 & 0.142 & 0.500 & 0.550 & 0.536 \\
$C^{\sim}_{\searrow}$ & $D_{equitoken}$ & 0.578 & 0.650 & 0.514 & 0.179 & 0.502 & 0.620 & 0.536 \\
$C_{A}$ & $D_{equitoken}$ & 0.577 & 0.634 & 0.547 & 0.184 & 0.492 & 0.625 & 0.528 \\
$(C*h)^{\sim}_{\nearrow}$ & $D_{equitoken}$ & 0.577 & 0.638 & 0.559 & 0.168 & 0.495 & 0.485 & 0.527 \\
$C_{rand}$ & $D_{equitoken}$ & 0.579 & 0.615 & 0.548 & 0.215 & 0.493 & 0.625 & 0.523 \\
$C_{source}$ & $D_{equitoken}$ & 0.577 & 0.609 & 0.480 & 0.244 & 0.499 & 0.615 & 0.522 \\
$C_{source}$ & $D_{stratified}$ & 0.577 & 0.593 & 0.479 & 0.286 & 0.518 & 0.570 & 0.522 \\
$C_{PPL}$ & $D_{equitoken}$ & 0.582 & 0.635 & 0.490 & 0.129 & 0.498 & 0.610 & 0.517 \\
$C_{MATTR}$ & $D_{equitoken}$ & 0.579 & 0.627 & 0.529 & 0.141 & 0.498 & 0.665 & 0.516 \\
$C^{\{50\}}_{\nearrow}$ & $D_{equitoken}$ & 0.579 & 0.592 & 0.535 & 0.228 & 0.503 & 0.495 & 0.511 \\
$(C*h)^{\sim}_{\searrow}$ & $D_{equitoken}$ & 0.578 & 0.610 & 0.542 & 0.136 & 0.502 & 0.520 & 0.505 \\
$E_{masked}$ & ext & \textbf{0.665} & 0.508 & 0.483 & \textbf{0.419} & 0.502 & \textbf{0.965} & 0.504 \\
$C^E_{\searrow}$ & $D_{stratified}$ & 0.576 & 0.577 & 0.535 & 0.242 & 0.500 & 0.555 & 0.504 \\
$C^E_{\nearrow}$ & $D_{2024}$ & 0.570 & 0.518 & 0.499 & 0.412 & 0.506 & 0.925 & 0.503 \\
$E_{causal}$ & ext & 0.654 & 0.514 & 0.449 & 0.412 & 0.502 & 0.770 & 0.502 \\
$E_{mixed}$ & ext & 0.660 & 0.505 & 0.459 & 0.414 & 0.500 & 0.780 & 0.498 \\
$C^E_{\searrow}$ & $D_{2024}$ & 0.577 & 0.586 & 0.486 & 0.152 & 0.512 & 0.635 & 0.493 \\
$C^E_{\searrow}$ & $D_{equitoken}$ & 0.576 & 0.587 & 0.507 & 0.148 & 0.502 & 0.590 & 0.492 \\
$C^E_{\nearrow}$ & $D_{stratified}$ & 0.573 & 0.562 & 0.510 & 0.201 & 0.505 & 0.640 & 0.488 \\
$C^E_{\nearrow}$ & $D_{equitoken}$ & 0.570 & 0.471 & 0.502 & 0.415 & 0.502 & 0.685 & 0.473 \\
\end{longtable}

  }
  \footnotesize{
\begin{longtable}{lllllllll}
\caption{Macro accuracy for RoBERTa models across tasks, per benchmark and overall. $E_\cdot$ denotes baseline models from the BabyLM challange, the fine-tuning evaluation pipeline fails for the $E_{gpt2}$ model.} \\
\toprule
Curriculum & Dataset & (Super) GLUE & blimp\_filtered & supplement\_filtered & entity\_tracking & ewok\_filtered & wug\_adj\_nominalization & Macro acc \\
\midrule
\endfirsthead
\caption*{Continued from previous page} \\
\toprule
Curriculum & Dataset & (Super) GLUE & blimp\_filtered & supplement\_filtered & entity\_tracking & ewok\_filtered & wug\_adj\_nominalization & Macro acc \\
\midrule
\endhead
\midrule
\multicolumn{9}{r}{Continued on next page} \\

\endfoot
\bottomrule
\endlastfoot
$(C*h)^{\sim}_{\nearrow}$ & $D_{stratified}$ & 0.650 & 0.694 & 0.535 & 0.307 & 0.507 & 0.570 & \textbf{0.592} \\
$C^{\{50\}}_{\nearrow}$ & $D_{2024}$ & 0.634 & 0.700 & 0.578 & 0.268 & 0.501 & 0.690 & 0.590 \\
$C_{source}$ & $D_{2024}$ & 0.635 & 0.683 & 0.563 & 0.290 & 0.507 & 0.670 & 0.583 \\
$C_{A}$ & $D_{stratified}$ & 0.629 & 0.698 & 0.555 & 0.238 & 0.497 & 0.460 & 0.579 \\
$C_\nearrow$ & $D_{2024}$ & 0.643 & 0.664 & 0.535 & 0.309 & 0.504 & 0.780 & 0.576 \\
$C^E_{\searrow}$ & $D_{2024}$ & 0.641 & 0.689 & 0.530 & 0.237 & 0.500 & 0.655 & 0.576 \\
$C_{MATTR}$ & $D_{2024}$ & 0.642 & 0.701 & 0.556 & 0.186 & 0.505 & 0.715 & 0.575 \\
$(C*h)^{\sim}_{\searrow}$ & $D_{2024}$ & 0.645 & \textbf{0.702} & 0.570 & 0.171 & 0.500 & 0.680 & 0.573 \\
$C_{A}$ & $D_{2024}$ & 0.636 & 0.691 & 0.529 & 0.187 & 0.505 & 0.675 & 0.568 \\
$C_{PPL}$ & $D_{2024}$ & 0.633 & 0.614 & 0.534 & 0.402 & 0.500 & 0.620 & 0.561 \\
$C^{\sim}_{\nearrow}$ & $D_{2024}$ & 0.639 & 0.658 & 0.567 & 0.239 & 0.501 & 0.715 & 0.559 \\
$C^{\{50\}}_{\nearrow}$ & $D_{stratified}$ & 0.647 & 0.690 & 0.571 & 0.126 & 0.503 & 0.650 & 0.558 \\
$C^{\sim}_{\searrow}$ & $D_{2024}$ & 0.637 & 0.692 & 0.561 & 0.117 & 0.502 & 0.775 & 0.557 \\
$(C*h)^{\sim}_{\searrow}$ & $D_{stratified}$ & 0.650 & 0.689 & 0.538 & 0.140 & 0.502 & 0.560 & 0.557 \\
$C^E_{\searrow}$ & $D_{stratified}$ & 0.647 & 0.680 & 0.563 & 0.162 & 0.499 & 0.530 & 0.556 \\
$C^{\sim}_{\searrow}$ & $D_{stratified}$ & 0.638 & 0.690 & 0.556 & 0.128 & 0.503 & 0.460 & 0.555 \\
$C_{\searrow}$ & $D_{stratified}$ & 0.643 & 0.679 & 0.559 & 0.149 & 0.500 & 0.610 & 0.554 \\
$C_{\searrow}$ & $D_{2024}$ & 0.639 & 0.675 & 0.565 & 0.149 & 0.501 & 0.765 & 0.553 \\
$(C*h)^{\sim}_{\nearrow}$ & $D_{2024}$ & 0.626 & 0.675 & 0.571 & 0.148 & \textbf{0.511} & 0.745 & 0.553 \\
$C_{MATTR}$ & $D_{stratified}$ & 0.648 & 0.683 & 0.535 & 0.127 & 0.502 & 0.505 & 0.551 \\
$C_{source}$ & $D_{stratified}$ & 0.636 & 0.668 & 0.560 & 0.177 & 0.499 & 0.560 & 0.551 \\
$E_{gpt2}$ & ext & nan & 0.673 & \textbf{0.591} & 0.189 & 0.498 & 0.390 & 0.551 \\
$C_\nearrow$ & $D_{stratified}$ & 0.644 & 0.677 & 0.530 & 0.125 & 0.499 & 0.715 & 0.548 \\
$C^{\sim}_{\nearrow}$ & $D_{stratified}$ & 0.638 & 0.677 & 0.535 & 0.123 & 0.502 & 0.540 & 0.546 \\
$C_\nearrow$ & $D_{equitoken}$ & 0.611 & 0.575 & 0.476 & 0.409 & 0.501 & 0.925 & 0.538 \\
$C^{\{50\}}_{\nearrow}$ & $D_{equitoken}$ & 0.605 & 0.575 & 0.488 & 0.409 & 0.500 & 0.690 & 0.536 \\
$C^E_{\searrow}$ & $D_{equitoken}$ & 0.600 & 0.573 & 0.485 & 0.406 & 0.502 & 0.720 & 0.534 \\
$C_{source}$ & $D_{equitoken}$ & 0.609 & 0.569 & 0.488 & 0.407 & 0.500 & 0.855 & 0.533 \\
$(C*h)^{\sim}_{\nearrow}$ & $D_{equitoken}$ & 0.612 & 0.570 & 0.471 & 0.409 & 0.498 & 0.690 & 0.532 \\
$C_{PPL}$ & $D_{equitoken}$ & 0.605 & 0.566 & 0.486 & 0.411 & 0.501 & 0.770 & 0.531 \\
$C_{\searrow}$ & $D_{equitoken}$ & 0.606 & 0.564 & 0.475 & 0.409 & 0.496 & 0.690 & 0.528 \\
$C^{\sim}_{\nearrow}$ & $D_{equitoken}$ & 0.602 & 0.558 & 0.484 & 0.411 & 0.500 & 0.720 & 0.525 \\
$C^{\sim}_{\searrow}$ & $D_{equitoken}$ & 0.612 & 0.559 & 0.450 & 0.409 & 0.491 & 0.600 & 0.523 \\
$(C*h)^{\sim}_{\searrow}$ & $D_{equitoken}$ & 0.614 & 0.557 & 0.484 & 0.406 & 0.501 & 0.495 & 0.523 \\
$C_{MATTR}$ & $D_{equitoken}$ & 0.603 & 0.552 & 0.455 & 0.409 & 0.499 & 0.525 & 0.518 \\
$C^E_{\nearrow}$ & $D_{stratified}$ & 0.594 & 0.550 & 0.462 & 0.414 & 0.492 & 0.400 & 0.516 \\
$C_{rand}$ & $D_{stratified}$ & 0.591 & 0.542 & 0.467 & 0.408 & 0.504 & 0.520 & 0.512 \\
$C^E_{\nearrow}$ & $D_{equitoken}$ & 0.605 & 0.531 & 0.506 & 0.411 & 0.501 & 0.795 & 0.512 \\
$C_{PPL}$ & $D_{stratified}$ & 0.645 & 0.535 & 0.488 & 0.411 & 0.497 & 0.195 & 0.510 \\
$E_{masked}$ & ext & \textbf{0.665} & 0.508 & 0.483 & \textbf{0.419} & 0.502 & \textbf{0.965} & 0.504 \\
$E_{causal}$ & ext & 0.654 & 0.514 & 0.449 & 0.412 & 0.502 & 0.770 & 0.502 \\
$E_{mixed}$ & ext & 0.660 & 0.505 & 0.459 & 0.414 & 0.500 & 0.780 & 0.498 \\
$C^E_{\nearrow}$ & $D_{2024}$ & 0.591 & 0.516 & 0.459 & 0.407 & 0.497 & 0.540 & 0.496 \\
$C_{rand}$ & $D_{equitoken}$ & 0.595 & 0.506 & 0.438 & 0.418 & 0.494 & 0.660 & 0.492 \\
$C_{A}$ & $D_{equitoken}$ & 0.597 & 0.501 & 0.441 & 0.413 & 0.489 & 0.510 & 0.487 \\
$C_{rand}$ & $D_{2024}$ & 0.601 & 0.462 & 0.465 & 0.409 & 0.507 & 0.490 & 0.466 \\
\end{longtable}

  }
\end{landscape}
\footnotesize{
\begin{longtable}{lllll}
\caption{Average \% $R^2$ gain for Llama models in the reading benchmarks (not included in the main paper). $E_\cdot$ denotes baseline models from the BabyLM challenge.} \\
\toprule
Curriculum & Dataset & Eye Tracking Score & Self-Paced Reading Score & Avg \\
\midrule
\endfirsthead
\caption*{Continued from previous page} \\
\toprule
Curriculum & Dataset & Eye Tracking Score & Self-Paced Reading Score & Avg \\
\midrule
\endhead
\midrule
\multicolumn{5}{r}{Continued on next page} \\
\endfoot
\bottomrule
\endlastfoot
$E_{causal}$ & ext & 0.102 & \textbf{0.029} & \textbf{0.065} \\
$E_{masked}$ & ext & \textbf{0.103} & 0.027 & 0.065 \\
$E_{mixed}$ & ext & 0.099 & 0.025 & 0.062 \\
$C^E_{\nearrow}$ & $D_{equitoken}$ & 0.024 & 0.009 & 0.016 \\
$C^E_{\nearrow}$ & $D_{2024}$ & 0.021 & 0.010 & 0.016 \\
$C_{source}$ & $D_{stratified}$ & 0.011 & 0.001 & 0.006 \\
$C^E_{\searrow}$ & $D_{stratified}$ & 0.012 & 0.000 & 0.006 \\
$C^E_{\searrow}$ & $D_{2024}$ & 0.009 & 0.001 & 0.005 \\
$C_{source}$ & $D_{2024}$ & 0.006 & 0.001 & 0.003 \\
$C^{\{50\}}_{\nearrow}$ & $D_{equitoken}$ & 0.006 & 0.000 & 0.003 \\
$C_{rand}$ & $D_{equitoken}$ & 0.005 & 0.001 & 0.003 \\
$C_{source}$ & $D_{equitoken}$ & 0.005 & 0.001 & 0.003 \\
$C_{rand}$ & $D_{2024}$ & 0.005 & 0.001 & 0.003 \\
$C_{\searrow}$ & $D_{stratified}$ & 0.006 & 0.001 & 0.003 \\
$C_{\searrow}$ & $D_{equitoken}$ & 0.005 & 0.000 & 0.003 \\
$C_{\searrow}$ & $D_{2024}$ & 0.005 & 0.000 & 0.003 \\
$C_{PPL}$ & $D_{equitoken}$ & 0.006 & 0.001 & 0.003 \\
$C_{MATTR}$ & $D_{equitoken}$ & 0.005 & 0.000 & 0.003 \\
$C^{\{50\}}_{\nearrow}$ & $D_{stratified}$ & 0.005 & 0.001 & 0.003 \\
$(C*h)^{\sim}_{\nearrow}$ & $D_{2024}$ & 0.003 & 0.002 & 0.003 \\
$C_\nearrow$ & $D_{stratified}$ & 0.005 & 0.000 & 0.003 \\
$(C*h)^{\sim}_{\nearrow}$ & $D_{stratified}$ & 0.005 & 0.001 & 0.003 \\
$(C*h)^{\sim}_{\searrow}$ & $D_{equitoken}$ & 0.007 & 0.000 & 0.003 \\
$C^{\sim}_{\searrow}$ & $D_{equitoken}$ & 0.006 & 0.000 & 0.003 \\
$C_{PPL}$ & $D_{stratified}$ & 0.003 & 0.000 & 0.002 \\
$(C*h)^{\sim}_{\searrow}$ & $D_{2024}$ & 0.004 & 0.001 & 0.002 \\
$(C*h)^{\sim}_{\searrow}$ & $D_{stratified}$ & 0.005 & 0.000 & 0.002 \\
$C_{rand}$ & $D_{stratified}$ & 0.004 & 0.000 & 0.002 \\
$C^E_{\searrow}$ & $D_{equitoken}$ & 0.004 & 0.000 & 0.002 \\
$C^{\sim}_{\nearrow}$ & $D_{2024}$ & 0.003 & 0.000 & 0.002 \\
$C^{\sim}_{\nearrow}$ & $D_{equitoken}$ & 0.004 & 0.000 & 0.002 \\
$C^{\sim}_{\nearrow}$ & $D_{stratified}$ & 0.005 & 0.000 & 0.002 \\
$C^{\sim}_{\searrow}$ & $D_{2024}$ & 0.004 & 0.000 & 0.002 \\
$C_{PPL}$ & $D_{2024}$ & 0.003 & 0.001 & 0.002 \\
$C^{\sim}_{\searrow}$ & $D_{stratified}$ & 0.004 & 0.000 & 0.002 \\
$C_{A}$ & $D_{stratified}$ & 0.003 & 0.001 & 0.002 \\
$C_{A}$ & $D_{equitoken}$ & 0.004 & 0.000 & 0.002 \\
$C_{A}$ & $D_{2024}$ & 0.005 & 0.000 & 0.002 \\
$(C*h)^{\sim}_{\nearrow}$ & $D_{equitoken}$ & 0.005 & 0.000 & 0.002 \\
$C_\nearrow$ & $D_{equitoken}$ & 0.004 & 0.000 & 0.002 \\
$C_\nearrow$ & $D_{2024}$ & 0.002 & 0.000 & 0.001 \\
$C_{MATTR}$ & $D_{stratified}$ & 0.002 & 0.000 & 0.001 \\
$C_{MATTR}$ & $D_{2024}$ & 0.003 & 0.000 & 0.001 \\
$C^E_{\nearrow}$ & $D_{stratified}$ & 0.002 & 0.000 & 0.001 \\
$E_{gpt2}$ & ext & 0.001 & 0.000 & 0.001 \\
$C^{\{50\}}_{\nearrow}$ & $D_{2024}$ & 0.001 & 0.002 & 0.001 \\
\end{longtable}

  }
  \footnotesize{
\begin{longtable}{lllll}
\caption{Average \% $R^2$ gain for RoBERTa models in the reading benchmarks (not included in the main paper). $E_\cdot$ denotes baseline models from the BabyLM challenge.} \\
\toprule
Curriculum & Dataset & Eye Tracking Score & Self-Paced Reading Score & Avg \\
\midrule
\endfirsthead
\caption*{Continued from previous page} \\
\toprule
Curriculum & Dataset & Eye Tracking Score & Self-Paced Reading Score & Avg \\
\midrule
\endhead
\midrule
\multicolumn{5}{r}{Continued on next page} \\
\endfoot
\bottomrule
\endlastfoot
$E_{causal}$ & ext & 0.102 & \textbf{0.029} & \textbf{0.065} \\
$E_{masked}$ & ext & \textbf{0.103} & 0.027 & 0.065 \\
$E_{mixed}$ & ext & 0.099 & 0.025 & 0.062 \\
$C^E_{\nearrow}$ & $D_{stratified}$ & 0.076 & 0.015 & 0.046 \\
$C^E_{\nearrow}$ & $D_{2024}$ & 0.074 & 0.014 & 0.044 \\
$C_{rand}$ & $D_{stratified}$ & 0.070 & 0.016 & 0.043 \\
$C_\nearrow$ & $D_{stratified}$ & 0.075 & 0.009 & 0.042 \\
$C_{PPL}$ & $D_{stratified}$ & 0.071 & 0.012 & 0.041 \\
$C_{rand}$ & $D_{2024}$ & 0.064 & 0.011 & 0.037 \\
$C_{PPL}$ & $D_{2024}$ & 0.060 & 0.007 & 0.033 \\
$C^{\sim}_{\nearrow}$ & $D_{stratified}$ & 0.051 & 0.007 & 0.029 \\
$C^E_{\nearrow}$ & $D_{equitoken}$ & 0.045 & 0.011 & 0.028 \\
$C_\nearrow$ & $D_{2024}$ & 0.050 & 0.006 & 0.028 \\
$C_{A}$ & $D_{equitoken}$ & 0.045 & 0.012 & 0.028 \\
$C_{rand}$ & $D_{equitoken}$ & 0.041 & 0.012 & 0.027 \\
$(C*h)^{\sim}_{\searrow}$ & $D_{stratified}$ & 0.046 & 0.005 & 0.026 \\
$C^E_{\searrow}$ & $D_{stratified}$ & 0.043 & 0.004 & 0.024 \\
$C^E_{\searrow}$ & $D_{2024}$ & 0.045 & 0.003 & 0.024 \\
$C^{\sim}_{\searrow}$ & $D_{2024}$ & 0.039 & 0.007 & 0.023 \\
$(C*h)^{\sim}_{\searrow}$ & $D_{2024}$ & 0.039 & 0.005 & 0.022 \\
$C_{source}$ & $D_{2024}$ & 0.039 & 0.003 & 0.021 \\
$C^{\sim}_{\nearrow}$ & $D_{2024}$ & 0.035 & 0.007 & 0.021 \\
$C_{A}$ & $D_{2024}$ & 0.036 & 0.005 & 0.021 \\
$C_{\searrow}$ & $D_{stratified}$ & 0.036 & 0.003 & 0.020 \\
$C^{\{50\}}_{\nearrow}$ & $D_{stratified}$ & 0.034 & 0.004 & 0.019 \\
$C_{A}$ & $D_{stratified}$ & 0.034 & 0.003 & 0.018 \\
$C^{\sim}_{\searrow}$ & $D_{stratified}$ & 0.033 & 0.003 & 0.018 \\
$C_{\searrow}$ & $D_{2024}$ & 0.030 & 0.005 & 0.017 \\
$C^{\{50\}}_{\nearrow}$ & $D_{2024}$ & 0.033 & 0.002 & 0.017 \\
$(C*h)^{\sim}_{\nearrow}$ & $D_{stratified}$ & 0.031 & 0.002 & 0.016 \\
$C_{MATTR}$ & $D_{2024}$ & 0.029 & 0.003 & 0.016 \\
$C_{source}$ & $D_{stratified}$ & 0.024 & 0.003 & 0.014 \\
$(C*h)^{\sim}_{\nearrow}$ & $D_{2024}$ & 0.019 & 0.001 & 0.010 \\
$C_{\searrow}$ & $D_{equitoken}$ & 0.015 & 0.003 & 0.009 \\
$(C*h)^{\sim}_{\searrow}$ & $D_{equitoken}$ & 0.015 & 0.003 & 0.009 \\
$C^{\sim}_{\nearrow}$ & $D_{equitoken}$ & 0.015 & 0.003 & 0.009 \\
$C_{source}$ & $D_{equitoken}$ & 0.016 & 0.003 & 0.009 \\
$C_{MATTR}$ & $D_{stratified}$ & 0.018 & 0.001 & 0.009 \\
$(C*h)^{\sim}_{\nearrow}$ & $D_{equitoken}$ & 0.014 & 0.003 & 0.008 \\
$C^{\{50\}}_{\nearrow}$ & $D_{equitoken}$ & 0.012 & 0.002 & 0.007 \\
$C_{PPL}$ & $D_{equitoken}$ & 0.011 & 0.003 & 0.007 \\
$C_\nearrow$ & $D_{equitoken}$ & 0.011 & 0.002 & 0.007 \\
$C^{\sim}_{\searrow}$ & $D_{equitoken}$ & 0.012 & 0.002 & 0.007 \\
$C^E_{\searrow}$ & $D_{equitoken}$ & 0.012 & 0.002 & 0.007 \\
$C_{MATTR}$ & $D_{equitoken}$ & 0.011 & 0.002 & 0.007 \\
$E_{gpt2}$ & ext & 0.001 & 0.000 & 0.001 \\
\end{longtable}

  }

    \begin{figure*}[t]
    \centering
    \includegraphics[width=0.99\linewidth]{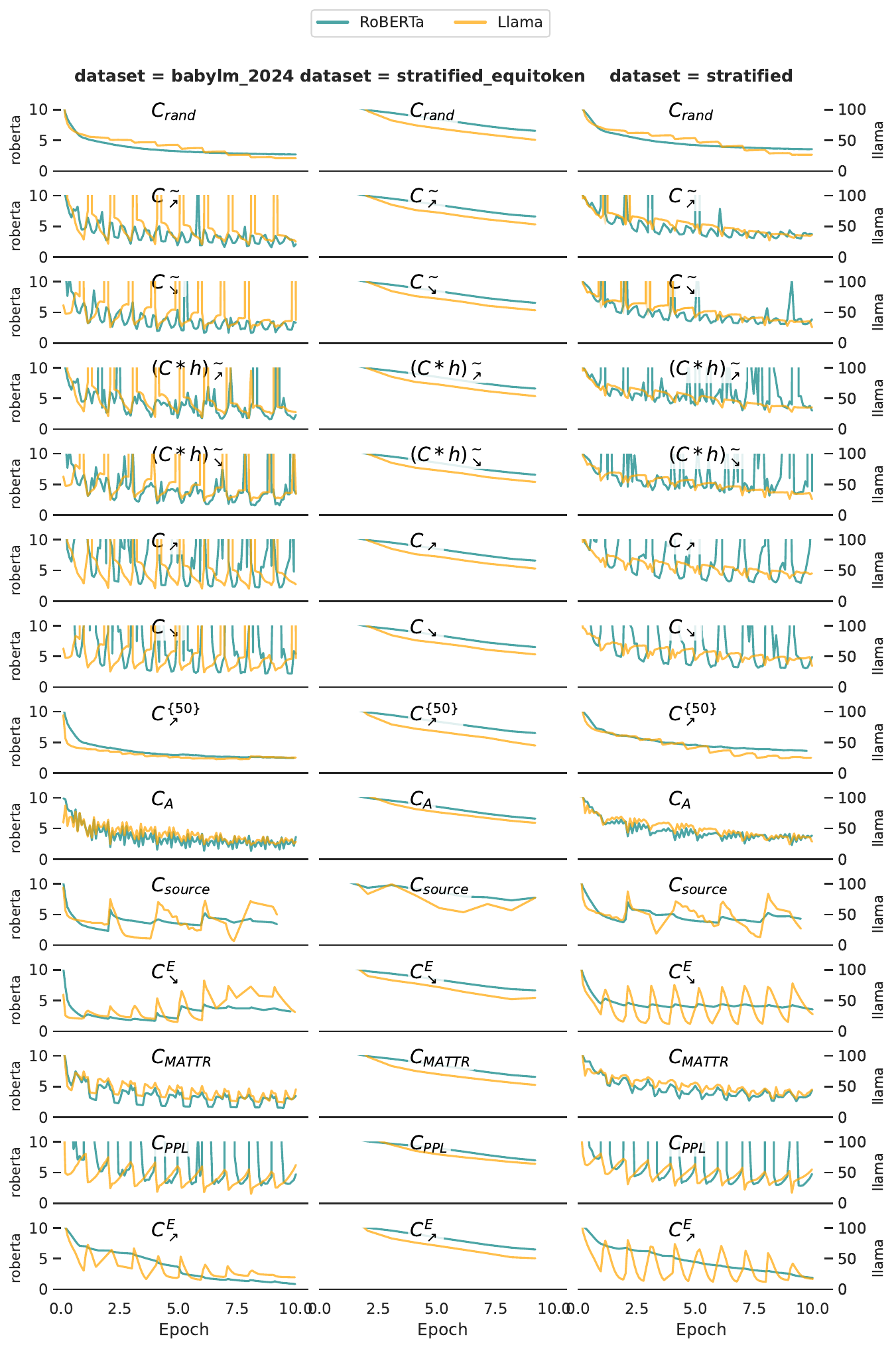}
    \caption{Training loss trajectories under different curricula.} 
    \label{train-loss-appendix}
\end{figure*}
\begin{figure*}[t]
    \centering
    \includegraphics[width=0.95\linewidth]{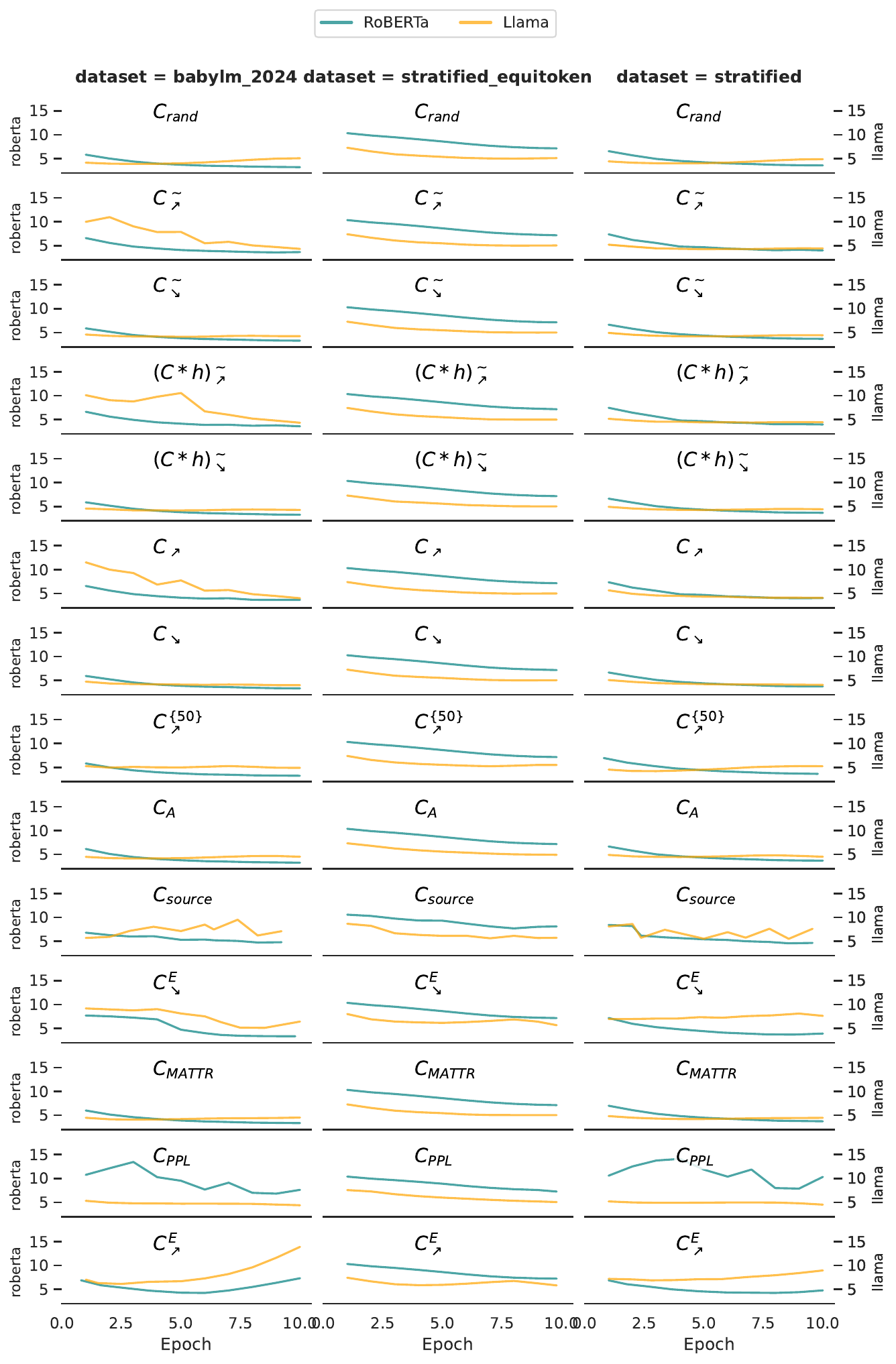}
    \caption{Evaluation loss trajectories under different curricula. We construct an evaluation set by sampling the 100M word 2024 BabyLM dataset ($D_{2024}$ is the 10M version; \citealp{choshen_call_2024}). $|D_{eval}| = 0.05 \cdot |D_{2024}|$.} 
    \label{eval-loss-appendix}
\end{figure*}
\twocolumn

\end{document}